\documentclass[10pt]{article} 
\usepackage[accepted]{tmlr}

\usepackage{amsmath,amsfonts,bm}

\def\eqref#1{equation~\ref{#1}}

\def\1{\bm{1}}

\DeclareMathAlphabet{\mathsfit}{\encodingdefault}{\sfdefault}{m}{sl}
\SetMathAlphabet{\mathsfit}{bold}{\encodingdefault}{\sfdefault}{bx}{n}

\def\gO{{\mathcal{O}}}

\newcommand{\E}{\mathbb{E}}

\newcommand{\R}{\mathbb{R}}

\usepackage{hyperref}
\usepackage{url}

\usepackage{times}
\usepackage{latexsym}

\usepackage[T1]{fontenc}

\usepackage[utf8]{inputenc}

\usepackage{microtype}

\usepackage{url}
\usepackage{multirow}
\usepackage{amsbsy}
\usepackage{times}
\usepackage{latexsym}
\usepackage{amsfonts}
\usepackage{amsthm}
\usepackage{dsfont}
\usepackage{amssymb}
\usepackage{amsmath}
\usepackage{pgfplots}
\pgfplotsset{compat=1.17}
\usepackage{booktabs}
\usepackage{diagbox}
\usepackage{graphicx}
\usepackage{subcaption}
\usepackage{xspace}
\usepackage{xcolor}
\usepackage{enumitem}
\newcolumntype{C}{>{\raggedright\arraybackslash}X}
\title{Learning the Transformer Kernel}

\author{Sankalan Pal Chowdhury \hfill spalchowd@inf.ethz.ch \\
Department of Computer Science\\
ETH Zürich
\AND
Adamos Solomou \hfill solomou.adamos@gmail.com\\
Department of Computer Science\\
ETH Zürich
\AND
Avinava Dubey \hfill avinavadubey@google.com\\
Google Research \\
Mountain View, CA
\AND
Mrinmaya Sachan \hfill 
mrinmaya.sachan@inf.ethz.ch\\
Department of Computer Science\\
ETH Zürich
}

\newcommand{\change}[1]{{\color{black}#1}}
\newcommand{\Change}[1]{{\color{black}#1}}

\def\month{07}  
\def\year{2022} 
\def\openreview{\url{https://openreview.net/forum?id=tLIBAEYjcv}} 

\def\gO{{\mathcal{O}}}
\newcommand{\bigO}[0]{\mathcal{O}}
\newcommand\mydots{\makebox[1em][c]{.\hfil.\hfil.}}

\newcommand{\Gen}{\textsc{Generative}}
\newcommand{\GenP}{\Gen-\prf}

\newcommand{\MG}{\GMM-\RKS}
\newcommand{\GMM}{\textsc{GMM}}
\newcommand{\PRF}{\GMM-\prf}
\newcommand{\FSGB}{\textsc{FastFood}}
\newcommand{\FRKS}{\FSGB-\RKS}
\newcommand{\FPRF}{\FSGB-\prf}
\newcommand{\GEN}{\Gen-\RKS}
\newcommand{\TF}{{Transformer}}
\newcommand{\LE}{{LinearElu}}
\newcommand{\PF}{{Performer}}
\newcommand{\attn}{\textsc{Attn}}
\newcommand{\bigbird}{{Big Bird}}
\newcommand{\reformer}{{Reformer}}
\newcommand{\longformer}{{Longformer}}
\newcommand{\linformer}{{Linformer}}
\newcommand{\sinkhorn}{{Sinkhorn}}
\newcommand{\synthesizer}{{Synthesizer}}
\newcommand{\kerneltransformer}{\change{\textsc{KL-Transformer}}}
\newcommand{\vanillatransformer}{Softmax Transformer}
\newcommand{\vanillatrans}{{Softmax Trans.}}
\newcommand{\RKS}{\textsc{RKS}}
\newcommand{\prf}{\textsc{PRF}}
\newcommand{\lra}{\textit{LRA}}
\newcommand{\listops}{\textit{ListOps}}
\newcommand{\txt}{\textit{Text}}
\newcommand{\image}{\textit{Image}}
\newcommand{\retrieval}{\textit{Retrieval}}

\newcommand{\glue}{\textit{GLUE}}

\usepackage{cleveref}
\Crefname{ALC@unique}{Line}{Lines}
\crefname{section}{\S}{\S\S}
\Crefname{section}{\S}{\S\S}
\crefformat{section}{\S#2#1#3}
\crefname{table}{Table}{Tables}
\crefname{figure}{Fig.}{Fig.}
\crefname{algorithm}{Alg}{Alg}
\crefname{algorithm}{Alg}{Alg}
\crefname{line}{line}{lines}
\crefname{appendix}{\S\!\!}{\S\!\!}
\crefname{thm}{Theorem}{}
\crefname{prop}{Prop.\@}{Props.\@}
\crefname{defin}{Definition}{Definitions}
\crefname{lemma}{Lemma}{Lemmata}
\crefname{cor}{Corollary}{Corollaries}
\crefname{equation}{}{}
\crefname{myexample}{Example}{Examples}

\usepackage{wrapfig}
\usepackage[textwidth=1.2in]{todonotes}
\newcommand*\iftodonotes{\if@todonotes@disabled\expandafter\@secondoftwo\else\expandafter\@firstoftwo\fi}  
\makeatother

\begin{document}

\maketitle

\begin{abstract}
In this work we introduce \kerneltransformer, a generic, scalable, data driven framework for learning the kernel function in \TF s.
Our framework approximates the \TF\ kernel as a dot product between spectral feature maps and learns the kernel by learning the spectral distribution. This not only helps in learning a generic kernel end-to-end, but also reduces the time and space complexity of \TF s from quadratic to linear. 
We show that \kerneltransformer s achieve performance comparable to existing efficient \TF\ architectures, both in terms of accuracy and computational efficiency. Our study also demonstrates that the choice of the kernel has a substantial impact on performance, and kernel learning variants are competitive alternatives to fixed kernel \TF s, both in long as well as short sequence tasks. 
\footnote{Our code and models are available at \url{https://github.com/cs1160701/OnLearningTheKernel}}
\end{abstract}

\section{Introduction}\label{sec:introduction}


\TF\ models \citep{vaswani2017transformers} have demonstrated impressive results on a variety of tasks dealing with language understanding~\citep{devlin2019bert, radford2018gpt, raffel2020text, brown2020gpt3}, image processing~\citep{parmar2018image, carion2020endtoend, lu2019ViLBERT}, as well as biomedical informatics~\citep{rives2020biological, ingraham2019protein, madani2020progen}.
Albeit powerful, due to the global receptive field of self-attention, the time and memory complexity of \vanillatransformer\ models scale quadratically with respect to the sequence length. As a result, the application of \TF s to domains with long contexts is rather limited. 
This limitation has spawned several efficient \TF\ designs \citep{liu2018memory, parmar2018image, child2019sparse, zaheer2020bigbird, beltagy2020longformer, roy2020routing, tay2020sinkhorn, kitaev2020reformer}. Kernelization offers one such design. The use of kernel feature maps allows to reformulate the computation of attention in a way that avoids the explicit computation of the full attention matrix which is the key bottleneck for \vanillatransformer. This also opens up new directions for more generic attention mechanisms.

\begin{wrapfigure}{r}{0.47\textwidth}
    \centering
    \includegraphics[width=0.47\textwidth]{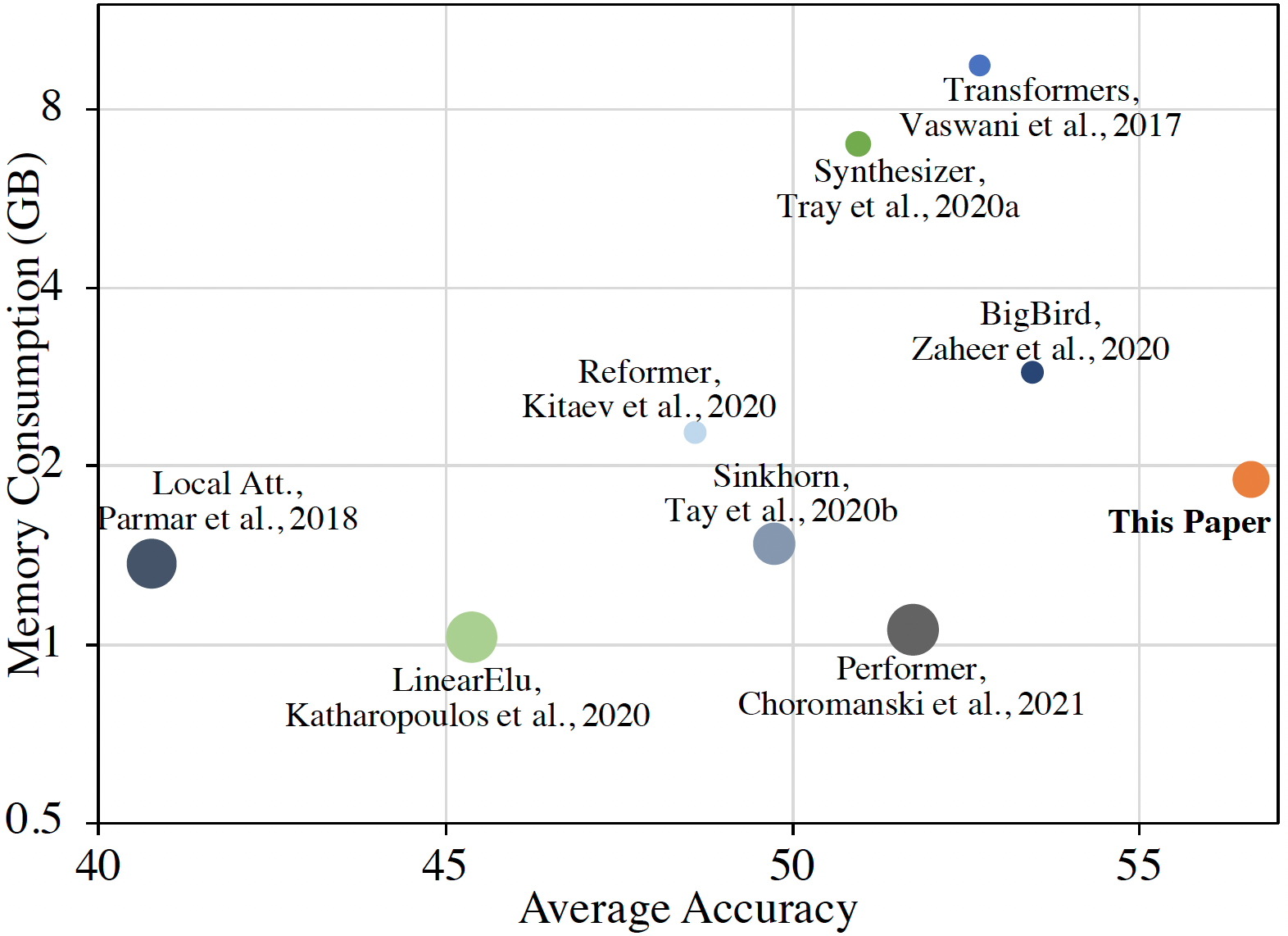}
    \caption{Peak memory (y-axis), average performance (x-axis) and speed (denoted by area of circle) for various efficient Transformer models (i.e. bigger circles in the bottom right corner are better) across three long sequence tasks ($>1024$ tokens) introduced in \lra\ \citep{tay2021lra}. All values except for ``This Paper'' are taken from \citet{tay2021lra}.\label{fig:lra-memory-accuracy-intro}}
    \vspace{-2mm}
\end{wrapfigure}

\citet{tsai2019dissection} first proposed a kernel-based formulation of the attention mechanism. However, the time and memory complexity of their approach remains quadratic with respect to the sequence length. To address this limitation, \citet{katharopoulos2020linear} expressed self-attention as the inner product of kernel feature maps and made use of the associative property to reduce the complexity from quadratic to linear. For their experiments, they used the arbitrarily chosen \LE\ feature map $f(x)=\max(x+1,e^x)$. \PF\ \citep{choromanski2021performers} replaces this with feature maps that can directly approximate the softmax kernel, thereby allowing the use of pre-trained \vanillatransformer\ weights in a linear time model. Concurrently with them, \citet{DBLP:journals/corr/abs-2103-02143} proposed a linear space and time method that added causal and recency based features to random Fourier methods. More recently, \citet{schlag2021linear} showed the formal equivalence of linearized self-attention mechanisms and fast weight programmers. While the aforementioned approaches provide a significant reduction in computational and memory requirements, this often comes at the cost of performance, as can be seen from Fig.~\ref{fig:lra-memory-accuracy-intro}. In this work, we posit that this is partly due to the fact that the similarity functions/kernels, including scaled-dot-product, were hand picked and not learnt from data. Thus, we explore whether kernel learning can help to bridge this gap in performance while retaining the scalability of efficient Transformers.

Although, to the best of our knowledge, kernel learning has never been explored within the framework of Transformers, kernel learning methods have been an ubiquitous tool in machine learning. 
The most notable among them is Random Kitchen Sinks \citep[\RKS;][]{rahimi2007random}, a data-independent framework for approximating shift-invariant kernels using an explicit feature map. In \RKS, the kernel is approximated by $\kappa(x,y) \approx \langle \phi(x),\phi(y)\rangle$, where the explicit feature map $\phi: \R^d \rightarrow \R^s$ is obtained by sampling from a spectral distribution defined by the inverse Fourier transform of the kernel function $\kappa$.
\citet{wilson2013gaussian} modeled the spectral density as a mixture of Gaussians, A la Carte \protect{\citep{yang2015alacarte}} proposed an optimization based framework for learning the spectral distribution, BaNK \citep{oliva2016bank} modeled the spectral distribution using an infinite mixture of Gaussians, while \citet{fang2020endtoend} implicitly modeled it using deep generative models. 
We build on these advances and incorporate them into the \TF\ framework.


\textbf{Contributions:} $\quad$ In this work, we propose \kerneltransformer, a scalable data driven framework for learning the kernel of Transformers and investigate whether a fully learnable kernel can help to improve the performance of linear, fixed kernel \TF s.
Thus, we 
introduce \TF s with learnable similarity functions, which happen to retain the linear complexity in terms of the sequence length.
We motivate our learning method using \RKS\ and learn the kernel by learning the corresponding spectral distribution.
In ~\Cref{sec:kl} we first propose to learn a generic Transformer kernel by explicitly approximating the spectral distribution using a Mixture of Gaussians (\GMM) and propose modifications to scale it further. In an attempt to further explore the trade off between computational complexity and accuracy we also propose to model the spectral frequency distribution of Transformer kernels implicitly by using deep generative models \citep{goodfellow2014generative}. 
Finally, we also propose a novel method to learn the spectral distribution of positive random feature (\prf) maps, which provides a better approximation of the softmax kernel \citep{choromanski2021performers}.

We analyse the expressivity and precision of our proposed models (\Cref{sec:analysis}) and show that the proposed  \GMM\ with positional encodings is Turing-complete~\citep{perez2018turing} with controllable variance.
%
We experimentally evaluate our models on \lra\ (tasks with long context), \glue\ (tasks with short context) and a synthetic dataset with controllable sparsity, and analyze the performance of our models (\Cref{sec:exp}, \Cref{sec:analysis}). In our experiments, we find that learnt kernels improve performance in long-context tasks, while staying competitive to the \vanillatransformer\ of the same size in short-context tasks. We also find that our models learn parameters that reduce the variance in their predictions, and can handle sparsity quite well. We also benchmark the computational efficiency of \kerneltransformer s 
and find that each of our proposed \kerneltransformer s  scales linearly with respect to the sequence length. We conclude with a short comparison between Random Kitchen Sinks (RKS, \citet{rahimi2007random}) and Positive Random Features \citet{choromanski2021performers} in terms of their performance and provide recommendations on which approach should be chosen under which circumstances.
\section{Kernel Learning in Transformers}
\label{sec:def}
We begin with the generalized formulation of self-attention proposed by \citet{tsai2019dissection}. Given a non-negative kernel function $\kappa(\cdot, \cdot): \R^{d_k} \times \R^{d_k} \rightarrow \R_+$, the output of the generalized self-attention mechanism at index $i$ operating on an input sequence $X = (x_1, ..., x_L) \in \R^{L\times d}$ is defined as
\begin{small}
\begin{equation}\label{eq:self-attention-kernel-quadratic}
    \attn(X)_i = \sum\limits_{j=1}^L\frac{\kappa(q_i, k_j)}{\sum_{j'=1}^L \kappa(q_i, k_{j'})}v_j.
\end{equation}
\end{small}
where $ k_i = x_i W^K, q_i = x_i W^Q, v_i = x_i W^V$ are linear transformations of the input sequence into keys, queries and values of dimension $d_q=d_k$ and $d_v$ respectively and $W^K \in \R ^{d \times d_k}$, $W^Q \in \R ^{d \times d_q}$, $W^V \in \R ^{d \times d_v}$.
While the formulation in Eq. (\ref{eq:self-attention-kernel-quadratic}) is more generic and defines a larger space of attention functions, it suffers from a quadratic time and memory complexity. To reduce the quadratic time and memory, we briefly review the method of random Fourier features for the approximation of kernels \citep{rahimi2007random}. The details of the method will help motivate and explain our models. 

\paragraph{Random Fourier Features for Kernels:}\label{sec:rff-kernel-learning}
At the heart of this method lies the theorem of Bochner~\citep{rudin1990bohner} which states that a continuous shift invariant kernel $\kappa(q,k) = \tilde{\kappa} (q-k)$ over arbitrary variables $q$ and $k$ is a positive definite function if and only if $\tilde{\kappa}(\delta)$ is the Fourier transform of a non-negative measure $\rho(\omega)$. Moreover, if $\tilde{\kappa}(0) = 1$, then Bochner's theorem guarantees that $\rho(\omega)$ is a normalized density function, i.e.
\begin{small}
\begin{align}\begin{split}
    \tilde{\kappa}(q-k) = \int_{\R^d} \rho(\omega) \exp\big(i \omega^T (q-k)\big) \,d\omega \label{eq:bohner-integral} = \E_{\omega \sim \rho}\big[\exp(i \omega ^T q)\exp(i \omega ^T k)^*\big].
\end{split}\end{align}
\end{small}
\citet{rahimi2007random} proposed to sample from the spectral density $\rho(\omega)$ for a Monte Carlo approximation to the integral in Eq. (\ref{eq:bohner-integral}). Specifically, for real valued kernels, they define $\kappa(q, k) \approx \phi(q)^T \phi(k)$, where $\omega_i \sim \rho(\omega)$ and
\begin{small}
\begin{align}\begin{split}
    \phi(x)& := RKS(x,\Omega=(\omega_1,\ldots,\omega_M)) := \frac{1}{\sqrt{M}}[\cos(\omega_1^Tx),\ldots,\cos(\omega_M^Tx), \sin(\omega_1^Tx),\ldots,\sin(\omega_M^Tx)] 
\end{split}\end{align} \label{eq:feature-map-definition}
\end{small}
To learn a kernel, we can either learn a parametric function $\kappa(\cdot,\cdot)$ or learn the corresponding parameterized feature map $\phi(\cdot)$ directly, which corresponds to learning the spectral density $\rho(\omega)$~\citep{wilson2013gaussian,yang2015alacarte,oliva2016bank}. 
In this paper, we focus on the latter because this helps us in keeping the computational complexity linear in the sequence length $L$. This can be achieved by rewriting Eq. (\ref{eq:self-attention-kernel-quadratic}) as
$
    \attn(X)_i = \frac{\phi(q_i)^T(\sum_{j=1}^L \phi(k_j) v_j^T)}{\phi(q_i)^T \sum_{j'=1}^L \phi(k_{j'})} 
$. To the best of our knowledge this is the first attempt to learn the kernel of the generalized self-attention mechanism (Eq. \ref{eq:self-attention-kernel-quadratic}).
\subsection{Learning Kernels in Spectral Domain}\label{sec:kl}

\paragraph{\MG:} 
Our objective is to enable learning of any translation invariant kernel. This is realizable if we can learn the spectral distribution.
Gaussian Mixture Models (\GMM s) are known universal approximators of densities and hence may approximate any spectral distribution. \GMM s have been shown to be useful for kernel learning for regression and classification tasks \citep{wilson2013gaussian,oliva2016bank}. Thus, to learn the kernel of the generalized self-attention mechanism (Eq.~\ref{eq:self-attention-kernel-quadratic}), we 
model the spectral distribution of the kernel as a parameterized \GMM, i.e.
\begin{align}
& \rho(\omega) = \sum_{c=1}^C \pi_c \mathcal{N}(\mu_c, \Sigma_c) \Leftrightarrow \label{eq:mixkern}
\kappa(q,k) = \sum_{c=1}^{C} \pi_c e^{(i \mu_c^T(q-k) - \frac{1}{2}(q-k)^T\Sigma_c(q-k))}
\end{align}
Here $\{\mu_c\in \R^d, \Sigma_c\in\R^{d^2}\}_{c=1}^C$ are the learnable parameters of the feature map and $C$ is the number of components in the Gaussian mixture.  It can be shown using  Plancherel’s Theorem that $\rho(\omega)$ can approximate any shift invariant kernel \citep{silverman1986density}.
Since we are working with only real valued kernels, the corresponding kernel reduces to $\kappa(q,k) = \sum_{c=1}^C \pi_c e^{(- \frac{1}{2}(q-k)^T\Sigma_c(q-k)))}   \cos{(\mu_c^T(q-k))}$. 

To speedup learning, we assume that $\pi_c = \frac{1}{C}$ and parameterize the feature map with spectral frequency, $\Omega =(\omega_{c,1},\ldots,\omega_{C,M})$  as:  
\begin{align}\begin{split}
   & \phi_{\MG}(x) := RKS(x,\Omega), \quad   \omega_{c,m} = \Sigma_c n_m + \mu_c, \quad n_m \sim \mathcal{N}(\mathbf{0},\mathbf{I}). 
    \label{eq:self-attention-mixgauss-linear}
\end{split}\end{align}
This allows us to sample $n_m\sim \mathcal{N}(\mathbf{0},\mathbf{I})$  and learn the parameters of the feature map, ($\{\mu_c\in \R^{d_q}, \Sigma_c\in\R^{d_q^2}\}_{c=1}^C$) end-to-end along with the other parameters of the Transformer.

\paragraph{\FRKS:}
\MG\ removes the quadratic dependency on context length, but we still need to calculate $\Omega^T\mathbf{Q}$ and $\Omega^T\mathbf{K}$ (where $\Omega=[\omega_1,\omega_2,\ldots,\omega_M]$) which takes $\mathcal{O}(Md_qL)$ time and $\mathcal{O}(Md_q+d_qL+ML)$ space, which can be too much if $M$ is large. For further gains in scalability, we  approximate the spectral frequency matrix $\Omega$, using the product of Hadamard matrices \citep[FastFood;][]{le2013fastfood}, such that the computation can be done in time log-linear in $M$, i.e.:
\begin{align}\begin{split}
    \phi_{\FRKS}(x) := RKS(x,V), \quad
    \textmd{where } V = \frac{1}{\sigma \sqrt{d_q}}SHG\Pi HB. \label{eq:fastfood}
\end{split}\end{align}
Here, $\Pi \in \{0,1\}^{d_q \times d_q}$ is a permutation matrix, $H$ is the Walsh-Hadamard matrix, $B$ is a diagonal random matrix with $\{\pm 1\}$ entries, $G$ is a diagonal matrix with Gaussian entries drawn from $\mathcal{N}(0,1)$ and finally $S$ is a random diagonal scaling matrix that  makes the row lengths non-uniform. The entire multiplication can be carried out in logarithmic time, and the space requirement is reduced by storing diagonal matrices instead of full matrices. For $M>d_q$ we use multiple blocks, and the only restriction is that we need $M|d_q$. In order to make  this learnable, \citet{yang2015alacarte} proposed making $S$ and optionally $G$ and $B$ learnable. For the main paper, we keep all three learnable (the case where only $S$ is learnable is discussed in Appendix \ref{Ablations}). 

\paragraph{\GEN:}
If we increase the number of components ($C$) in \MG, the computation and space complexity increases dramatically. Instead, to learn a more generic kernel, without blowing up the computational complexity, we use deep generative models (DGMs). DGMs have achieved impressive results in density estimation \citep{goodfellow2014generative, kingma2014autoencoding,richardson2018gans, ruthotto2021generative} and end-to-end kernel learning in the spectral domain for classification~\citep{fang2020endtoend}.

\begin{figure}[t]
    \centering
    \includegraphics[width=0.75\textwidth]{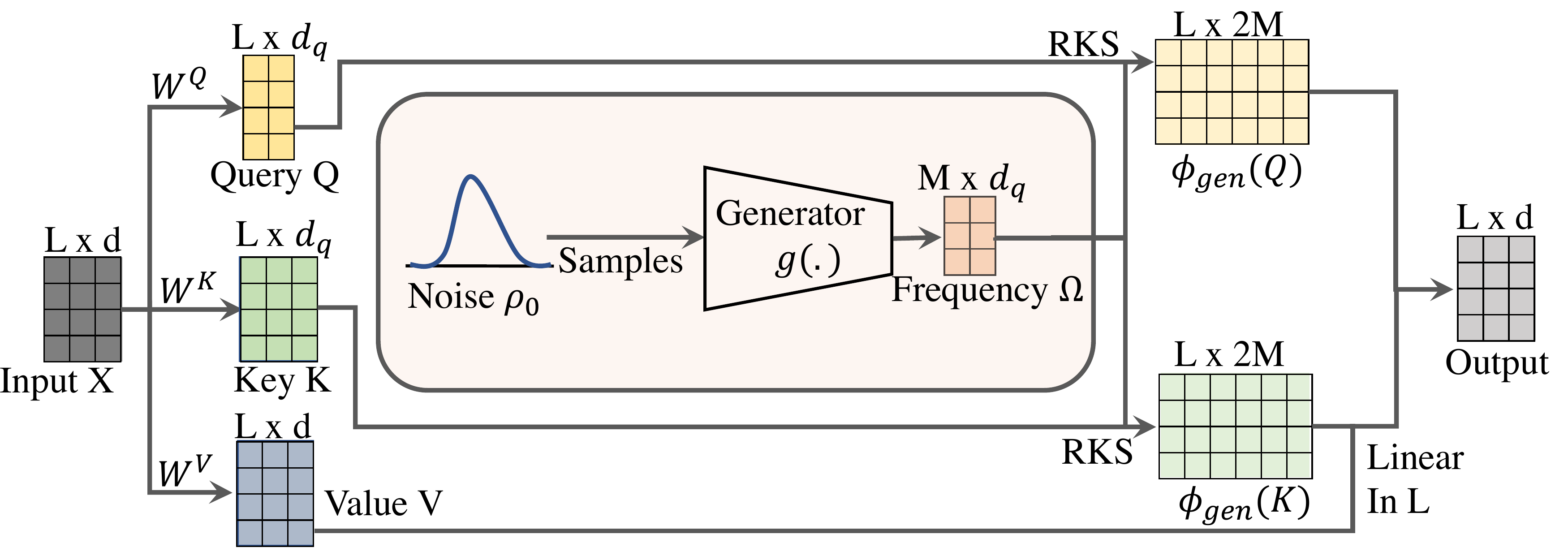}
    \caption{Generalized self-attention with deep generative RKS. $Q, V$ and $K$ are linear transformations of input, $X$. The generator generates spectral frequency ($\Omega$) from an implicit spectral distribution. Using RKS (Eq. \ref{eq:feature-map-definition}) we create the feature map $\phi_{gen}$ (Eq. \ref{eq:generative-rks}). The numerator of the output is calculated as $\phi_{gen}(Q)(\phi_{gen}(K)V)$ while the denominator is $\phi_{gen}(q_i)^T \sum_{j'=1}^L \phi_{gen}(k_{j'})$ making  attention linear in sequence length $L$. 
    }
    \label{fig:grff}
\end{figure}

\change{In \GEN\, we use a DGM to replace the Gaussian probability distribution from \MG\ with an arbitrary probability distribution. This DGM acts as a generator similar to the ones used by variational auto-encoders used in computer vision \citep{kingma2014autoencoding}. In particular, to make the sampling process differentiable, we use the reparameterization trick, where a learnable neural network (called the generator, and denoted by g in Figure \ref{fig:grff}) transforms samples from a simple noise distribution ($\rho_0$ in Fig \ref{fig:grff}) into samples from the desired distribution.
\begin{align}
    \omega_m = g(n_m),\quad n_m \sim \rho_o(\cdot) 
\end{align}
The generator network is trained end to end with the whole model, allowing it to choose the best possible distribution for the given data.
These samples ($\Omega=[\omega_1,\omega_2,\ldots,\omega_M]$ in Fig \ref{fig:grff}) are then used in Random Kitchen Sinks as follows:}
\begin{align}
    &\phi_{\GEN}(x) := RKS(x,\Omega),\label{eq:generative-rks}\quad \omega_m \sim g(\rho_o)
\end{align}
We experimented with various configurations and eventually and chose to learn a generator network which consisted of $4$ fully connected layers with batch normalisation and LeakyReLU activation, followed by a single fully connected layer with $\tanh$ activation.  \change{While this methodology allows us to generalise the gaussian distribution to a much larger class of distributions, it also causes a blowup in the number of parameters e.g. a $4+1$ layer constant width generator, as used by us would require $5d^2+5d$ parameters as opposed to the $d^2+d$ parameters in \MG.
To counter this and to improve generalization, we share the same generator network across all heads in a layer, which means that the different heads only differ in the Query/Key/Value projection matrix.}

\paragraph{Positive Random Features (\prf):}
Until now, we focused on feature maps defined using \RKS. While our formulation is very general, recently it was shown that positive random features provide a better approximation to both Gaussian and Softmax kernels (see Lemma 2 in \citealt{choromanski2021performers}). 
In particular they showed that $\kappa(q,k) = \exp(q^Tk) = \E_{\omega \sim \mathcal{N}(0,I)}[\exp(\omega^T q - \frac{\| q \|^2}{2}) \exp(\omega^T k - \frac{\|k\|^2}{2} )]$
and demonstrated that Monte Carlo approximation to this expectation leads to a low variance estimate of the softmax kernel. 
Moreover, the presence of only positive values within the randomized feature map ensures that kernel estimates remain strictly non-negative. 
To incorporate this prior knowledge,
we propose a novel kernel learning framework in which we learn the spectral density while using the feature map corresponding to the above expectation.  
For instance, when we model the spectral distribution of $\Omega=(\omega_1,\mydots, \omega_M)$ using \GMM\ ($\psi = \sum_{c=1}^C \pi_c \mathcal{N}(\mu_c, \Sigma_c)$) we have that:

\begin{align}
   & \kappa(q,k) := \E_{\omega \sim \psi}[e^{(\omega^T q - \| q \|^2)} e^{(\omega^T k - \|k\|^2)}] \label{eq:prf-def}, \quad PRF(x,\Omega) := \frac{e^{-\|x\|^2}}{\sqrt{M}}[e^{\omega_{1}^T x},\mydots, e^{\omega_{M}^T x} ] \\
    & \phi_{\PRF}(x) := PRF(x,(\omega_{c,1},\mydots,\omega_{C,M})), \quad
    \omega_{c,m} = \Sigma_c n_m + \mu_c, \quad n_m\sim \mathcal{N}(\mathbf{0},\mathbf{I}) \label{eq:fgmm}
\end{align}

Similarly we redefine the other two methods as:
\begin{align}
& \phi_{\GenP}(x) := PRF(x,\Omega),\quad
 \omega_m = g(n_m), \quad  n_m \sim \rho_o(\cdot) \label{eq:fgen}\\
 & \phi_{\FPRF}(x) := PRF(x,V),\label{eq:fff} \quad \textmd{where } V = \frac{1}{\sigma \sqrt{d_q}}SHG\Pi HB.
\end{align}
 To the best of our knowledge we are the first to explore kernel learning with positive random features. 

\begin{table}[t]
\centering
\resizebox{0.75\columnwidth}{!}{
\begin{tabular}{@{}l l l l l@{}}
    \toprule
  \textbf{ Model} & & \textbf{Space Complexity} & & \textbf{Time Complexity}\\
    \cmidrule{1-1} \cmidrule{3-3} \cmidrule{5-5} 
    \vanillatransformer   & &     $\gO(L^2(1+ d_q/L))$                                                   & & $\gO(L^2d_q)$                                                   \\
\PF & & $\gO(L(d_q+ M+Md_q/L))$                                            & & $\gO(LMd_q)$                                                          \\
\LE    & & $\gO(L(d_q +d_q^2/L))$                                                   & & $\gO(Ld_q^2)$                                                              \\
\midrule
\MG                           & & $\gO(L(d_q + C(d_q^2/L + Md_q/L+M)))$                                    & & $\gO(MC(d_q^2+Ld_q))$                                                   \\
\PRF       &  & $\gO(L(d_q+ CMd_q/L+CM))$                                           & & $\gO(LCMd_q)$     \\  
\textit{\FSGB(\RKS/\prf)}   & & $\gO(L(d_q + M + Md_q/L))$                                              & & $\gO(LMd_q)$                                                    \\
\textit{\Gen(\RKS/\prf)}      &  & $\gO(L(d_q + d_q^2/L +Md_q/L+M))$                                  &  & $\gO(M(d_q^2+Ld_q))$                                                   \\
 
     \bottomrule
\end{tabular}
}
\caption{Space and time complexity of self-attention kernel of \kerneltransformer s compared with \vanillatransformer~\citep{vaswani2017transformers}, \PF~\citep{choromanski2021performers}, and \LE~\citep{katharopoulos2020linear}. $L$ refers to length of context, $d_q$=$d_v$ is the query/key/value dimension, while $M$ is the number of samples (where applicable).} 
\label{tbl:complexity}
\end{table}

\subsection{Analysis}\label{sec:analysis}
In this section, we explore what can be said about the expressivity of the proposed linear time \kerneltransformer s. While our understanding of the properties of \TF s is rudimentary, we would still like to know whether the known properties extend to our models. For example, it has been shown that  \vanillatransformer s  and its sparse counterparts are Turing complete \citep{perez2018turing,zaheer2020bigbird}.
This raises the question as to whether the proposed linear \kerneltransformer s are also Turing complete?

It is easy to see that the generalized kernel self-attention (Eq. \ref{eq:self-attention-kernel-quadratic}) includes the softmax kernel and hence should satisfy the properties of \vanillatransformer. Interestingly, we can also show that this property holds for \MG\ \TF s with number of components $C=1$, (for a more systematic definition, see Section \ref{Definitions}). More formally,
\paragraph{Theorem 1:}\textit{The class of \MG\ \TF s with positional embeddings is Turing complete.}

Proof is in the Appendix \ref{sec:Proof}.\ We also show that:
\paragraph{Theorem 2:}\textit{The class of \PRF\ \TF s with positional embeddings is Turing complete.}

\noindent For a detailed proof, see Appendix \ref{sec:Proof}.\\
\\
 Since the sampling of $\omega$ in Equations \ref{eq:self-attention-mixgauss-linear} and \ref{eq:prf-def} is random, we have some stochasticity in the attention layers of our model. We now show that the Mean Square Error (MSE)  of the estimation can be reduced by reducing the eigenvalues of the learnt covariance matrix. In particular, we show that:
\paragraph{Theorem 3:}\textit{Let $\mu$ and $\Sigma=S^TS$ be the learnt mean an covariance matrices of the sampling distribution. Further let $q$ and $k$ be the parameters, and $p=k-q$ and $o=k+q$ be their vector difference and sum respectively, and m be the number of samples. The MSE of the linear approximations is given by:}
\begin{align}
& MSE_{\MG} =\frac{2}{m}\cos^2(\mu^Tp)(1-e^{-||
S^Tp||^2})^2\\
& MSE_{\PRF}=\frac{1}{m}e^{-2(||q||^2+||k||^2-\mu^To)}(e^{2||S^To||}-e^{||S^To||})
\end{align}
It is interesting to note that in  this formulation, $MSE_{\MG}$\footnote{For the proof of Theorem 3, we use the secific case of $C=2$, $\mu_1+\mu_2=0$ and $\Sigma_1=\Sigma_2$ which is used in the experiments. Since Theorem 1 and 2 however, hold in general} is bounded above by $\frac{2}{m}$ while no such bound can be established for $MSE_{\PRF}$. For the detailed proof see Supplementary Section \ref{sec:MSE} 
\paragraph{Complexity Analysis:}
While all of our models have linear complexity with respect to the context length $L$, differences still exist amongst the various methods. Notable, \MG\ and \Gen\ have quadratic time and space complexity in the query size $d_q$. Both the \FSGB\  methods avoid this approximation, whereas \PRF\ avoids this by the use of a diagonal covariance matrix. The complexities are listed in Table \ref{tbl:complexity}.\\\\
Another factor that controls timing is the sampling of $\Omega$. Sampling too frequently can lead to significant slowdowns whereas sampling too few times can lead to biased learning. For our experiments, we resample every $100$ training iterations, although this can be changed. A detailed list of all hyperparameters along with implementation details are provided in Appendix~\ref{sec:appen:expt}.
\section{Experiments}\label{sec:exp}

\subsection{Does kernel learning improve performance of fixed kernel methods on longer sequences?}\label{sec:exp-lra}

Long Range Arena (\lra; \citealt{tay2021lra}) is a diverse benchmark for the purpose of evaluating the ability of sequence models to reason under long-context scenarios. It includes tasks that vary both in terms of the context length (ranging from $1K$ to $4K$ tokens) as well as the data modalities (including text and mathematical expressions). We evaluate the \kerneltransformer\ architectures introduced in Section \ref{sec:kl} on the \textit{Text}, \textit{Retrieval} and \textit{ListOps} tasks from \lra\ which deal with sentiment prediction from IMDB Reviews, document similarity classification and pre-order arithmetic calculations respectively. Both \textit{Text} and \textit{Retrieval} use character level encodings, bringing their maximum length to $4000$ tokens. The \textit{ListOps} dataset is synthetically generated and has a fixed length of $2000$ tokens. For more details on the tasks, we refer the interested reader to the original paper by \citealt{tay2021lra}.
    
    

\paragraph{Setup:} To ensure a fair comparison, we closely follow the same data preprocessing, data split, model size and training procedure as in~\citep{tay2021lra}. Within each task, a common configuration is used across all \kerneltransformer\ models based on the configuration specified in the \lra\ code repository\footnote{\url{https://github.com/google-research/long-range-arena}}. We outline the hyperparameters for all tasks in Table \ref{table:lra-hyperparameters} in the Appendix. 

\begin{table}[!t]
\begin{center}
    
  \resizebox{0.7\columnwidth}{!}{%
  \begin{tabular}{@{}l|c|ccc|r@{}}
    \toprule
    \textbf{Model} & Complexity & \listops & \txt & \retrieval & \textbf{Avg.}\\
     & & 2K & 4K & 4K \\
    \midrule\midrule

    Random Predictor  & NA                          & 10.00 & 50.00 & 50.00 & 36.67 \\
    \midrule
    \multicolumn{6}{c}{Baseline Models}  \\
    \midrule
    \vanillatrans\ (\citeauthor{vaswani2017transformers}) & $\gO(L^2)$ & 36.38 & 64.27 & 57.46 & 52.70 \\
    \synthesizer\ (\citeauthor{tay2021synthesizer}) & $\gO(L^2)$ & 36.50 & 61.68 & 54.67 & 50.95\\
    \sinkhorn\ (\citeauthor{tay2020sinkhorn}) & $\gO((L/B)^2) $ & 34.20 & 61.20 & 53.83 & 49.74\\
    Sparse Trans. (\citeauthor{child2019sparse}) & $\gO(L\sqrt{L})$      & 35.78 & 63.58 & 59.59 & 52.98 \\
    \reformer\ (\citeauthor{kitaev2020reformer}) & $\gO(L\log{L})$  & 36.30 & 56.10 & 53.40 & 48.60\\
    Local Attention (\citeauthor{parmar2018image}) & $\gO(LK)$  & 15.95 & 52.98 & 53.39 & 40.77\\
    \longformer\ (\citeauthor{beltagy2020longformer}) & $\gO(LK)$ & 36.03 & 62.85 & 56.89 & 51.92\\
    \linformer\ (\citeauthor{wang2020linformer}) & $\gO(L)$ & 35.49 & 53.49 & 52.27 & 52.56\\
    \bigbird\ (\citeauthor{zaheer2020bigbird}) & $\gO(LK)$ & 37.08 & 64.02 & 59.29 & 53.46\\
    \LE\ (\citeauthor{katharopoulos2020linear})  & $\gO(L)$    & 17.15 & 65.90 & 53.09 & 45.38\\
    \PF\ (\citeauthor{choromanski2021performers}) & $\gO(L)$ & 36.00 & 65.40 & 53.82 & 51.74\\
    \midrule
    \multicolumn{5}{c}{Kernelized Transformers}  \\
    \midrule
    \MG\ (Eq. \ref{eq:self-attention-mixgauss-linear}) &$\gO(L)$   & 18.55 & 63.95 & 58.64 & 47.05 \\
    \FRKS\  (Eq. \ref{eq:fastfood}) & $\gO(L)$ & 18.65 & \underline{65.67} & 61.92 & 48.75\\
    \GEN\ (Eq. \ref{eq:generative-rks})     & $\gO(L)$ & 18.50 & \textbf{66.50} & 64.76 & 49.92 \\
    \PRF\  (Eqs. \ref{eq:prf-def}, \ref{eq:fgmm} )  & $\gO(L)$ & 36.96 & 62.64 & 65.27 & 54.96\\
    \FPRF\   (Eqs. \ref{eq:prf-def}, \ref{eq:fff} ) & $\gO(L)$ & \underline{37.05} & 64.66 & \textbf{71.13} & \textbf{57.61}  \\
    \GenP\ (Eqs. \ref{eq:prf-def}, \ref{eq:fgen} ) & $\gO(L)$ & \textbf{37.42} & 62.90 & {69.81} & \underline{56.71} \\
    \bottomrule
  \end{tabular}
  }
\end{center}
  \caption{Experimental results on the \lra\ benchmark. We report accuracy on the test set, except for \textit{Text} where validation set is used. The best model is in boldface and the second best is underlined if within 1\% f the best. Accuracy scores for all baseline models are from ~\citet{tay2021lra}. Here, $L$ refers to the sequence length, $K$ refers to the size of a local window and $B \ll L$ is a model specific parameter. For our models, accuracy is averaged over $100$ runs.}
  \label{table:lra-results} 
\end{table}

\paragraph{Results:} The results across all \lra\ tasks are summarized in Table \ref{table:lra-results}. \kerneltransformer\ variants that learn the kernel function directly from the data in an end-to-end manner outperform the baseline models by occupying both best and second-best performances. We find that \kerneltransformer s based on PRFs tend to outperform their \RKS\ counterparts which is also reflected on the average \lra\ score, with \FPRF\ being the best-performing model.  

\begin{figure}
    \centering 
        \includegraphics[width=0.48\columnwidth]{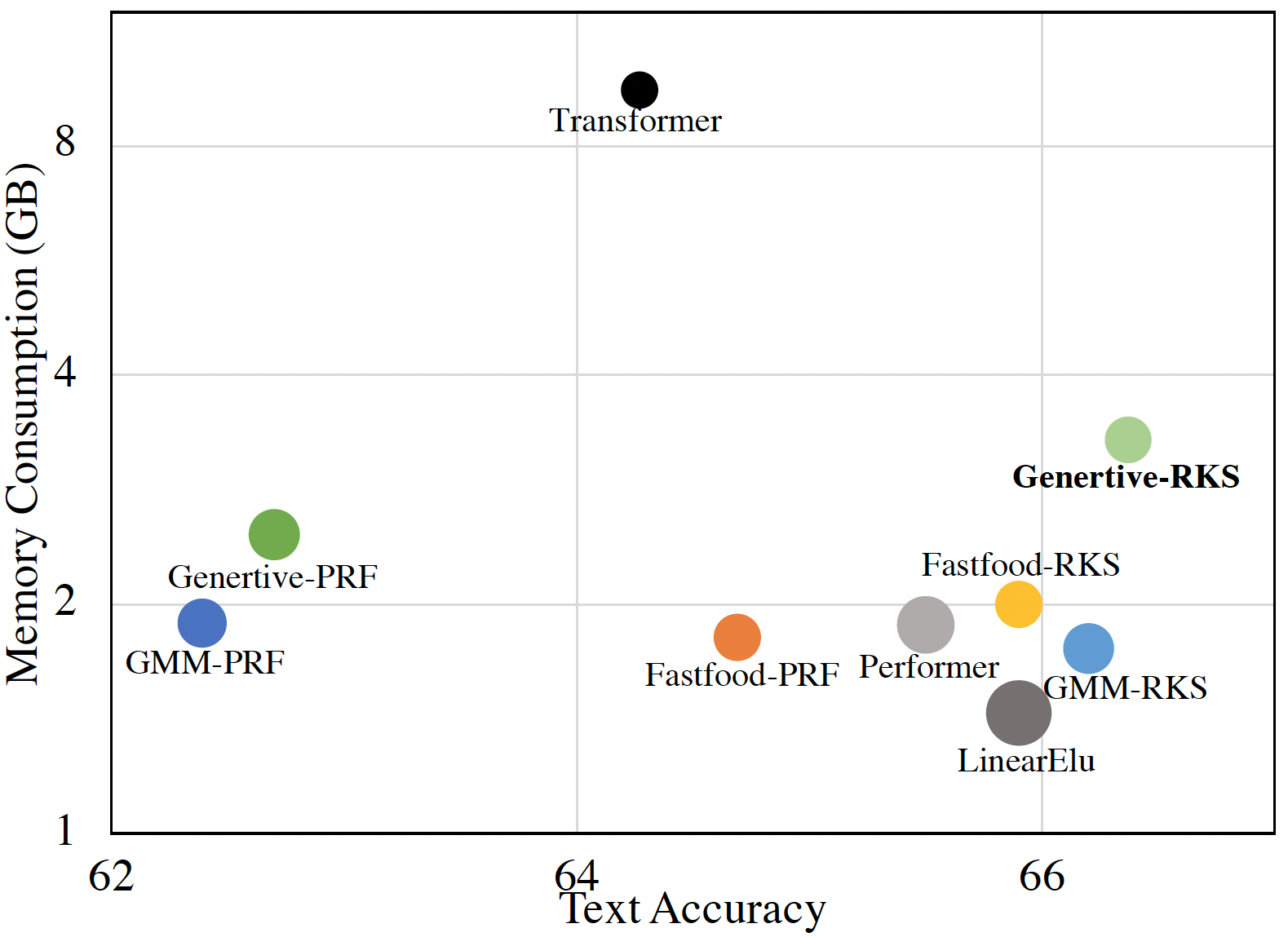}
        \includegraphics[width=0.48\columnwidth]{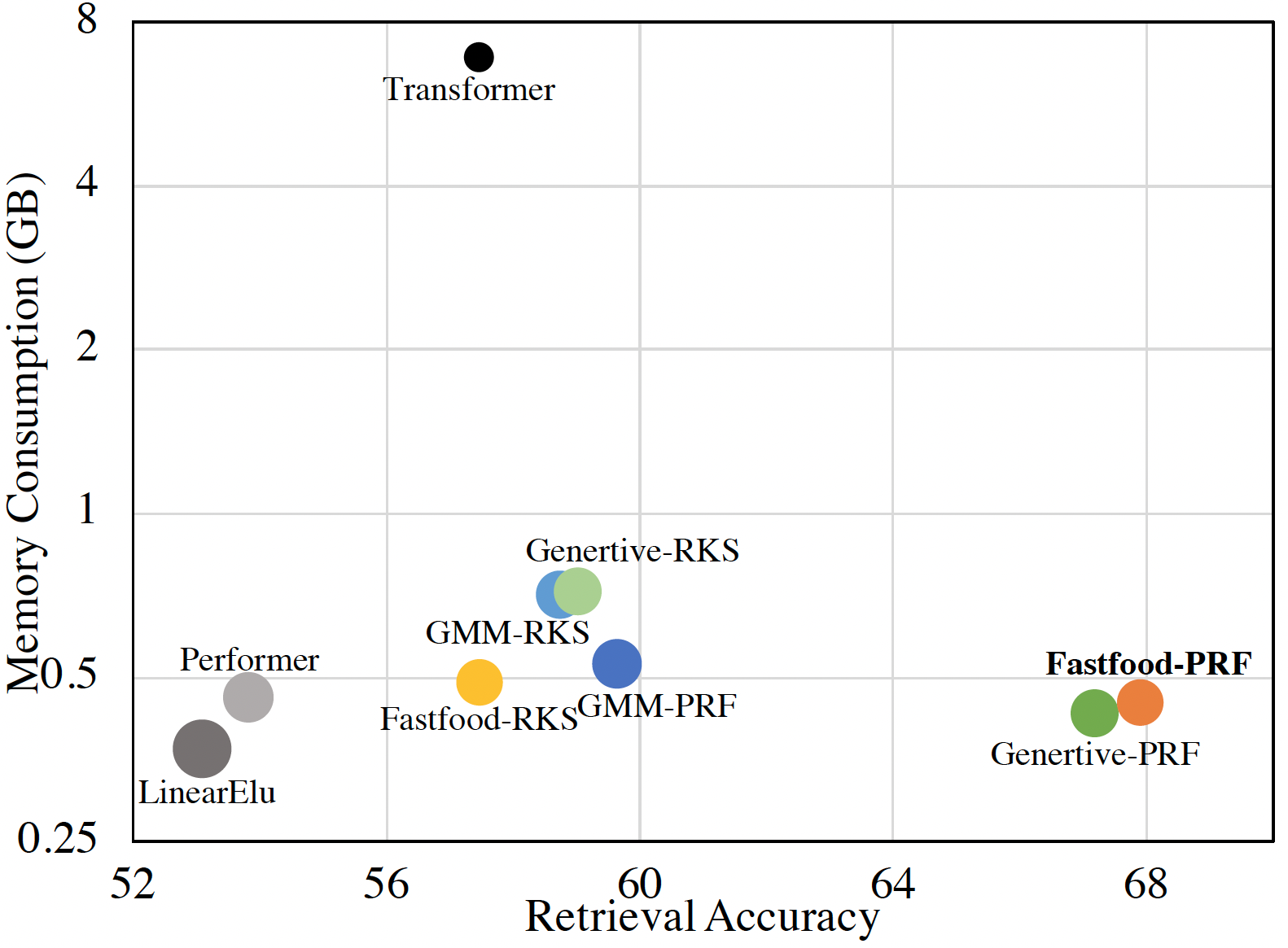}
    \caption{We compare the peak memory consumption (y-axis), performance (x-axis) and speed (denoted by area of circle) for the various \kerneltransformer\ architectures on the two \lra\ tasks with sequence length equal to $4K$. Memory usage refers to average memory usage across GPUs and speed (steps per second) for each model is reported relative to the speed of \vanillatransformer\ (larger circles denote faster models). For a similar graph on the \textit{ListOps} Task see Fig \ref{fig:lra-memory-accuracy-appendix} in the Supplementary} 
    \label{fig:lra-memory-accuracy}
\end{figure}

\subsection{Trade-off between Accuracy and Efficiency}\label{sec:exp-complexity}
We benchmark the efficiency of each \kerneltransformer\ in terms of peak memory usage and training speed and compare it against three baseline models from the \lra\ benchmark. Specifically, we compare against other efficient \TF\ architectures that employ fixed kernel feature maps (e.g. \LE\ and \PF) as well as the \vanillatransformer\ which is one of the strongest baseline models (see Table \ref{table:lra-results}). We conduct efficiency benchmarks on the two \lra\ tasks with sequence length equal to $4K$ in order to assess the efficiency of these methods in modelling tasks that require a long context (results for the other two datasets are included in the Appendix). Speed measurements (steps per second) refer to wall-clock training time (including overheads).  In both cases experiments are conducted on $8$ NVIDIA TITAN RTX GPUs. 
The comparison is illustrated in Figure \ref{fig:lra-memory-accuracy}. On the \txt\ task, \GEN\ is the best performing model, although it consumes more memory than the remaining \kerneltransformer\ architectures (it is still more efficient than the \vanillatransformer). \LE\ consumes the least amount of memory, while \MG\ provides a trade-off between the two. In \retrieval\ the situation is much clearer, with \FPRF\ and \GenP\ outperforming significantly other models in terms of accuracy while having very low memory consumption. The training speed  of \kerneltransformer s is of the same order of magnitude as Performers (as indicated by the area of each circle in Figure \ref{fig:lra-memory-accuracy}). 


\begin{wrapfigure}{r}{0.42\textwidth}
    \centering
    \vspace{-7mm}
    \includegraphics[width=0.42\columnwidth]{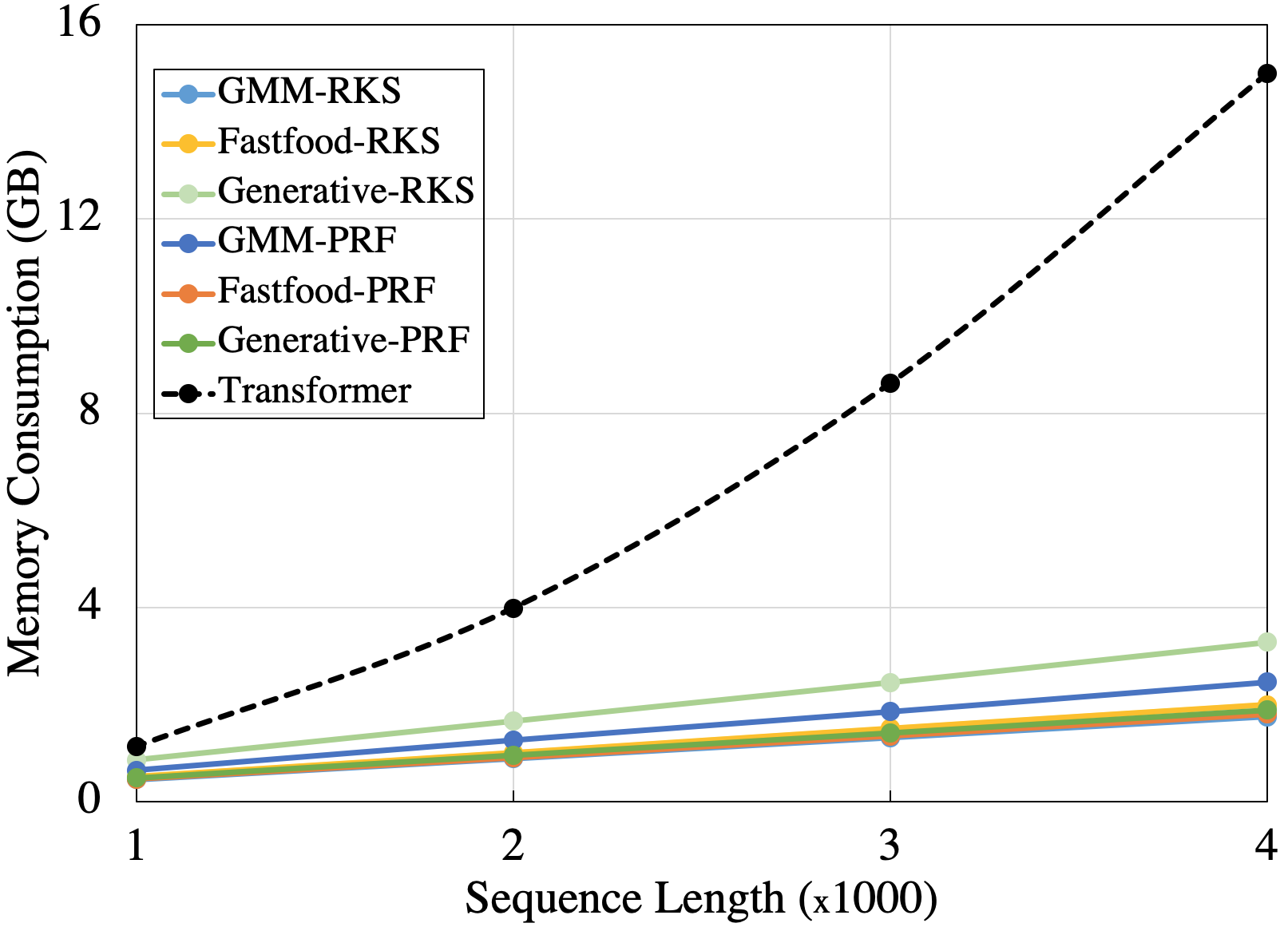}
    \vspace{-3mm}
    \caption{Memory vs sequence length\label{fig:lra-memory-length}} \vspace{-6mm}
\end{wrapfigure}
Lastly, in Figure \ref{fig:lra-memory-length}, we report the peak memory consumption as the sequence length changes from $1K$ to $4K$ on the \txt\ dataset. As expected, all our models have a linear increase in memory consumption with increasing sequence length, as opposed to the \vanillatransformer\ which has
dramatic increase in memory consumption. 
Furthermore, Figure \ref{fig:lra-memory-length_diff_data} in the Appendix reports the memory usage of each \kerneltransformer\ across all datasets. We find that \FPRF\ and \GenP\ are not only our best performing models on average, but they also consume the least memory among various {\kerneltransformer}s across all datasets. Thus, among the models proposed in this paper, we can recommend \FPRF\ and \GenP\ as the model that achieves the best accuracy with the least memory consumption.


\subsection{How do \kerneltransformer s perform on short sequence tasks? }\label{sec:exp-glue}

We compare the \kerneltransformer s and \vanillatransformer\ in a common transfer learning setting. We adopt the setting of BERT-like models \citep{devlin2019bert, liu2019roberta}, except that we have fewer layers and heads (see Table \ref{tab:glue-params} for details) and pre-train all models (including \vanillatransformer) on the WikiText-103 dataset \citep{merity2016pointer} using non-contextual WordPiece embeddings \citep{wu2016wordpiece}. Pre-trained models are then fine-tuned on the General Language Understanding Evaluation (\glue) benchmark \citep{wang2018glue}, a collection of resources for training and evaluating language understanding systems. All tasks in \glue\ consist of either one or two sentences as input and can therefore be easily captured with a context of size $512$. Since the context length is rather short, the difference between training and inference time across the various models is minimal. Instead, the main goal of this task is to assess how do \kerneltransformer s compare against \vanillatransformer s on a set of tasks where the later have been established as the \textit{de-facto} architecture. 

\begin{table}[!t]
  \centering
  \resizebox{0.8\columnwidth}{!}{
  \begin{tabular}{@{}l p{9mm} p{13mm} p{13mm} p{13mm} p{7mm} p{7mm} p{7mm} >{\raggedleft\arraybackslash}p{1cm}@{}}
    \toprule
    \textbf{Model} & \textbf{SST2} \par (acc) & \textbf{MRPC} \par (acc/f1) & \textbf{QQP} \par (acc/f1) & \textbf{MNLI-m/mm} \par (acc/acc) & \textbf{QNLI} \par (acc) & \textbf{WNLI} \par (acc) & \textbf{RTE} \par (acc) & \textbf{CoLA} \par (MCor)  \\
     \midrule
    \vanillatrans     & 0.81 & 0.70/0.82 & 0.83/0.76 & 0.64/0.64 & 0.68 & 0.56 & 0.6 & 0.18\\
    \midrule
    \FRKS    & 0.83 & 0.71/0.82 & 0.81/0.74 & 0.57/0.57 & 0.64 & 0.59 & 0.56 & 0.13\\
    \MG         & 0.80 & 0.70/0.82 & 0.77/0.69 & 0.47/0.48 & 0.60 & 0.61 & 0.57 & 0.07\\
    \GEN      & 0.81 & 0.70/0.82 & 0.81/0.73 & 0.59/0.58 & 0.63 & 0.62 & 0.58 & 0.16\\
 \FPRF      & 0.81 & 0.71/0.82 & 0.81/0.74 & 0.56/0.57 & 0.64 & 0.59 & 0.58 & 0.12\\
 \GenP      & 0.80 & 0.71/0.82 & 0.80/0.74 & 0.56/0.56 & 0.61 & 0.60 & 0.55 & 0.10\\
    \PRF         & 0.82 & 0.71/0.82 & 0.81/0.74 & 0.56/0.56 & 0.64 & 0.59 & 0.59 & 0.21 \\
    \bottomrule
  \end{tabular}
  }
  \caption{Results on the GLUE benchmark after fine-tuning on respective tasks. \kerneltransformer s continue to be competitive to \TF s even in short context problems.}
  \label{tab:GLUE}
\end{table}

The results on all downstream \glue\ tasks are shown in Table \ref{tab:GLUE}. Crucially, we demonstrate that there is no significant loss in performance compared to \vanillatransformer s when kernelized variants are used on short language modelling tasks. As illustrated in Table \ref{tab:GLUE}, \kerneltransformer s perform on par with the \vanillatransformer.

\section{Empirical Analysis}

In section \ref{sec:exp}, we observed that our models compare favourably with other linear models, with Positive Random Features (PRF) based models usually doing better that the ones based on Random Kitchen Sinks (RKS) (although RKS does better on the \textit{Text} task). In this section, we want to compare these linear kernels in terms of their empirical variance and how well they deal with sparsity. In Theorem $3$, we already noted that the MSE for both \MG\ and \PRF\ models decreases with decrease in eigenvalues of the covariance matrices, making it useful to learn such a matrix. We also noted that the variance in the output of \MG\ is bounded, and thus it should not face 
issues with 
sparse datasets when approximating the softmax kernel as show in \cite{choromanski2021performers}).
We test for these results empirically in subsections \ref{sec-compvar} and \ref{sec-sparse} respectively. We then conclude in \ref{sec-choice} with a discussion on which of our proposed models is best to use in a given scenario.
\subsection{Comparison of variance}
\label{sec-compvar}
\begin{table}[!t]
  \centering
  \resizebox{0.8\columnwidth}{!}{
\begin{tabular}{@{}l l r r c l l r r@{}}
\toprule
Task                       & Model               & Max Egv. & Mean Egv. & & Task                       & Model               & Max Egv. & Mean Egv. \\ \cmidrule{1-4} \cmidrule{6-9}
\multirow{2}{*}{Text}      & \MG  & 0.486          & 0.031  &   & \multirow{2}{*}{Retrieval} & \MG  & 0.499          & 0.053       \\
                           & \PRF & 0.096          & 0.025  &  &                            & \PRF & 0.186          & 0.042             \\ \bottomrule
\end{tabular}}
\caption{Distribution of Eigenvalues for \MG\ and \PRF\ models for the \textit{Text} and \textit{Retrieval} tasks. For a head by head distribution see Tables \ref{tab:egv}, \ref{tab:egvprf}, \ref{tab:egv-ret-mg} and \ref{tab:egv-ret-prf} }
\label{tab:eig}
\vspace{-4mm}
\end{table}

Looking at the eigenvalues of the covariance matrices of our final trained models\footnote{We only report eigenvalues for the GMM models since covariance matrices are not well defined in the Generator and FastFood. Also, the \textit{ListOps} task is left out of the entire analysis because its multi-class nature makes the analysis complex} in Table \ref{tab:eig}, we find that \MG\ and \PRF\ models trained on the \textit{Text} and \textit{Retrieval} tasks indeed learn to have eigenvalues significantly smaller than $1$ (which would be the case for non-learnable co-variances). 

While lower eigenvalues reduce the variance in our models significantly, it does not get completely nullified. To understand how much stochastic remains in our models, and also to compare RKS with PRF in regards to their variance, we look at the final outputs\footnote{While the theory makes a claim about the output of each attention head, evaluating every head at every layer would give us a large number of values to analyse. The output is considered before aplying the final sigmoid} the models produce when repeatedly run on the same example. We record the output produced by the model for $100$ runs on each datapoint, and calculate the following $3$ metrics to quantify the variance:
 

\begin{enumerate}
    \item \textbf{Relative Standard Deviation (RSD):} RSD is a standardised measure of  dispersion (see \cite{CurrieSvehla+1994+595+608}, def 3.9), defined as the ratio of the standard deviation of a set of observations to the absolute value of their mean. The ratio ensures that the measure is invariant to scaling (eg. multiplying the penultimate layer of the model by a constant factor does not affect RSD) while the absolute of the mean ensures that opposing classes don't cancel each other out. 
    \item \textbf{Prediction Inconsistency (PI):} While RSD quantifies the variance in the model's continuous output, in practice, 
    only the sign of the output is of interest as that decides the predicted class. As a way to quantify stochasticity in the discrete output, we count the number of times the output does not have the majority label, i.e., if the output is positive $x$ times, then we have $PI=\min{(x,100-x)}$. Alternately, it can be seen as the failure rate if we treat the majority prediction as the true class.
    
    \item \textbf{Accuracy Gain with Voting (AGV):} Since the final output has a propensity to change its sign, we can get a more robust prediction at inference time by running the model multiple times and considering the class it predicts more times as its output, instead of taking the prediction from a single inference step. Doing so, we are likely to get closer to the mean prediction (by the Law of Large Numbers, see \cite{revesz2014laws}), and we get accuracy scores which are slightly higher than those reported in Table \ref{table:lra-results}, and we call this value the \textit{Voting Accuracy (VA)}. AGV is then defined as the ratio between the voting accuracy and the regular accuracy. 
\end{enumerate}
\begin{figure}[th]
    \centering\vspace{-5mm}
    \includegraphics[width=\textwidth]{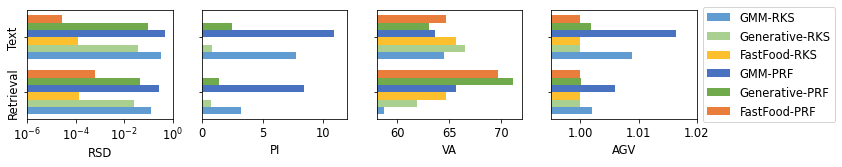}
    \caption{Values of variance metrics Relative Standard Deviation (RSD), Prediction Inconsistence(PI), Average Gain with Voting (AGV) as well as Voting Accuracy (VA) on the \textit{Text} and \textit{Retrieval} tasks.}
    \label{fig:analysis-bar}
\end{figure}

The above metrics, in addition to the \textit{Voting Accuracy}, are calculated over all instances in the validation and test set for \textit{Text} and \textit{Retrieval} tasks respectively, and plotted in Fig. \ref{fig:analysis-bar}. We note that our 3 metrics are all highly correlated with each other ($R^2$  values $0.95$, $0.98$ and $0.89$ for RSD-AGV, RSD-PI and AGV-PI respectively, see supplementary Fig \ref{fig:analysis-corr}, \ref{fig:analysis-corr-ret}). 

We notice that our RKS based models, in general, have lower variance as compared to our PRF models. 
We also note that Generator
and FastFood are able to further reduce the variance in the output,
possibly due to additional learnable parameters in Generator and
fewer sources of randomness in FastFood. We also notice that all models have greater variance in the Text task, and it is
possible that this \Change{is related to the better performance of} RKS based models in this task, and their inherent lower variance helps them
outperform their PRF counterparts. 

\subsection{Effectiveness on Sparse Datasets\label{sec-sparse}}


\cite{choromanski2021performers} demonstrated that if Random Kitchen Sinks are used to estimate the softmax-kernel, the MSE tends to $\infty$ as the raw softmax scores tend to $0$. This makes it particularly hard to deal with sparse datasets where multiple positions need near-zero attention. 

\begin{wrapfigure}{r}{0.45\textwidth}
    \centering\vspace{-9mm}
    \includegraphics[width=0.45\textwidth]{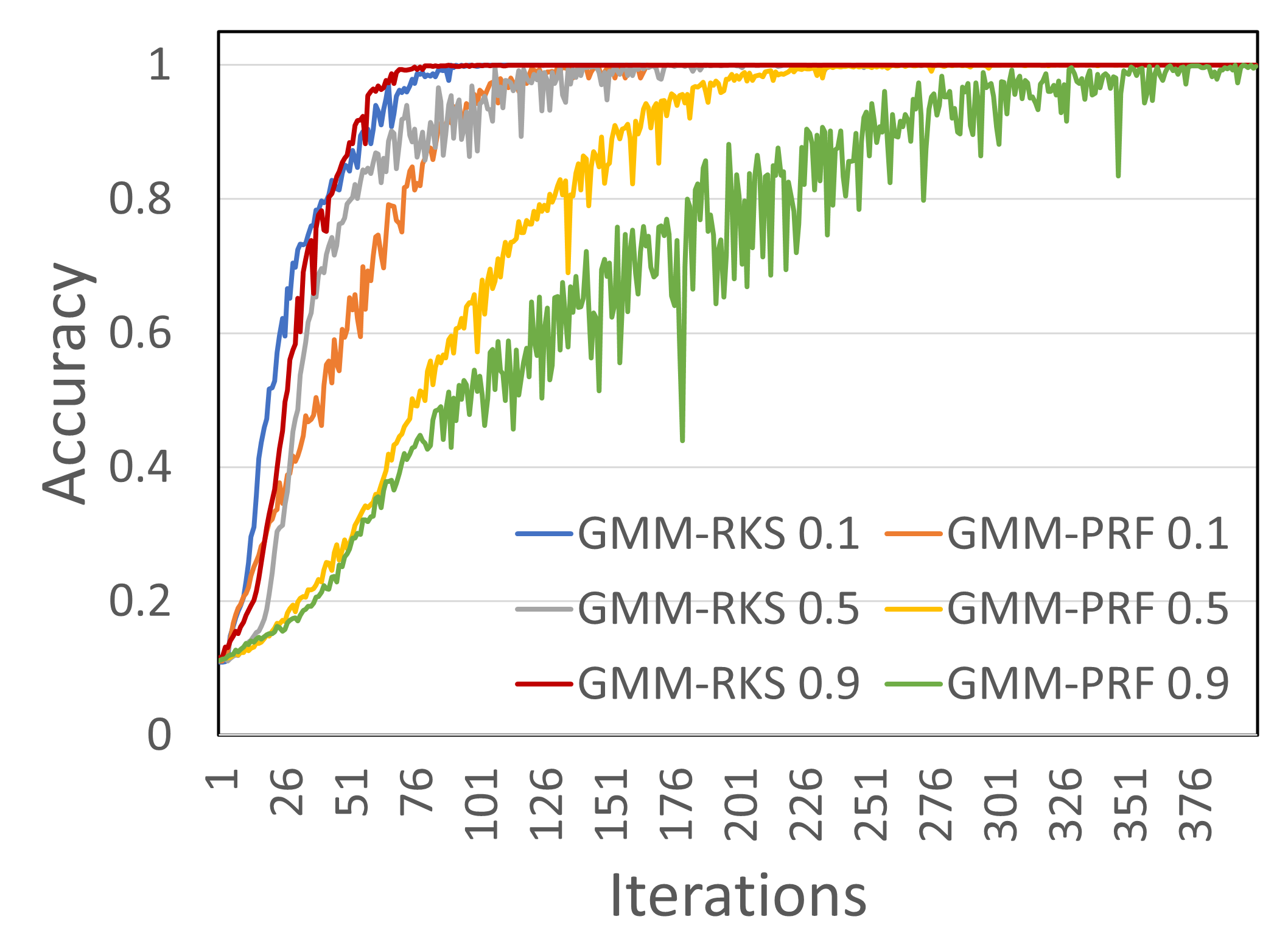} \vspace{-6mm}
    \caption{Learning curves for our synthetic experiment. The number after the model name is inversely proportional to sparsity in the dataset} \label{fig:synth}
    \vspace{-6mm}
\end{wrapfigure}
Thus, in order to test how sparsity in the dataset affects our models, we design a synthetic experiment where the model is fed a sequence of ordered pairs, where the first element, which we call \textit{score}, takes a value of $-1$ or $1$, and the second element, which we call \textit{relevance}, takes a value of $0$ or $1$. Looking at the sequence, the model must predict the sum of the scores at all positions with relevance $1$. Note that any position with relevance $=0$ does not contribute to the final answer, and therefore, does not need to be attended to. We construct the inputs by sampling the relevance of each position from Bernoulli$(p)$. Thus, the sparsity can be controlled by changing $p$ (for a detailed description of the task, see Supplementary).

Figure \ref{fig:synth} shows the learning curves for various sparsity levels in our synthetic experiment. Due to the simple nature of the task, all models eventually reach 100\% accuracy. We observe that the convergence time of \MG\ remains more or less unchanged with increasing sparsity, while for \PRF\ the model converges slower as sparsity decreases. We believe that the slower convergence of \PRF\ is \Change{correlated} to the variance in its output, which leads to variance in gradients. Observing the gradients propagated to the classifier layer (See Table \ref{tab:spar-grad} in Supplementary), we find that this is indeed the case, and not only is the variance in the gradients higher for \PRF, but also the mean is higher, making it take bigger steps in an uncertain direction.
This result provides us an insight into which models to use under which circumstances.

\subsection{Which Model to Use?}
\label{sec-choice}
PRF outperforms RKS is terms of accuracy in most of our experiments, especially if we are able to run multiple inference steps. Therefore, in general, we recommend using PRF over RKS. However, if consistency of predictions is important, or rapid training is required on a not-so-sparse task, one may consider RKS as well.

Finally, we observe that Generator and Fastfood methods always outperform the vanilla linearisations. Between them, Fastfood can be significantly faster if properly implemented on GPU. However, generators may perform better if a large amount of data is available since they provide a greater amount of flexibility.

\section{Related Work}\label{sec:related-work}

\subsection{Efficient Transformers}
A wide variety of approaches belong to the class of efficient \TF s \citep{tay2020survey}. We survey them below:

\change{\noindent\textbf{Sparse models:}} Memory Compressed Transformer \citep{liu2018memory} uses a convolution kernel (of size $K$) to sub-sample keys and values reducing the complexity to $\bigO(L^2K^{-1})$. Inspired by the notion of sparsity, \citet{child2019sparse} introduced sparse factorizations of the attention matrix to reduce the overall complexity to $\bigO(L \sqrt{L})$. Subsequently, \citet{roy2020routing} proposed the Routing Transformer which employs $K$-means clustering to learn dynamic sparse attention regions, achieving an overall complexity of $\bigO(L \sqrt{L})$. \change{Recently, \citet{sun2022sparse} proposed yet another sparse $\bigO(L \sqrt{L})$ approach that learns which bucket each query/key is to be placed in based on final attention values. Sparsity has also been achieved by efficiently subsampling queries and/or keys~\citep{chen2021sketching}}. \citet{kitaev2020reformer} proposed the \reformer, which reduces complexity to $\bigO(L\log L)$ by using locality-sensitive hashing to group together similar symbols. \citet{ye2019bptransformer} also proposed a $\bigO(L\log L)$ algorithm using binary partitions of data. There also exist other works that mainly focus on memory reduction~\change{\citep{liu2018memory, tay2020sinkhorn,https://doi.org/10.48550/arxiv.2106.06899}}. While these methods are faster than \vanillatransformer s, their asymptotic time complexity remains quadratic. \change{Further approaches attempt to minimize the constants involved in the quadratic attention, but keep the same assymptotic complexity~\citep{NEURIPS2021_2bd388f7, https://doi.org/10.48550/arxiv.2203.00091}.}

\change{\noindent\textbf{Local or Global Attention:}} \citet{parmar2018image} was one of the first local attention models that achieved $\bigO(L)$ complexity in both time and space, by using local attention over a constant length-context. \change{A similar local attention based method was also utilized by \citet{pmlr-v139-zhang21h} and \citet{https://doi.org/10.48550/arxiv.2203.12276}, the latter of which added special tokens at the start of each local attention block, which attend globally.  Another set of methods proposed to approximate the global attention by replacing the softmax to allow changing the order of matrix multiplication, making the calculation of the $QKV$ product linear in the context length~\citep{https://doi.org/10.48550/arxiv.2111.15588, https://doi.org/10.48550/arxiv.2202.10447,qin2022cosformer}. Linearised global attention is also used by \PF\ \citep{choromanski2021performers}, which has been built upon by other works. \citet{https://doi.org/10.48550/arxiv.2109.02377} and \citet{https://doi.org/10.48550/arxiv.2106.12566} both try to improve the performance of \PF\ by incorporating relative positional embeddings which they show makes it strictly stronger in terms of representability. \citet{DBLP:journals/corr/abs-2110-04367} tries to combine \RKS\ and \prf\ attentions by a learnable weight parameter to get the best of both worlds. Both these methods can be applied as is to our work. \citet{pmlr-v139-schlag21a} take a different view on this, claiming that having stochasticity hampers performance, and propose a deterministic feature map to avoid this.}

\change{\noindent\textbf{Multiple Types of Attention:}} \citet{beltagy2020longformer} proposed a $\bigO(L)$ method that combines the above two approaches by using local sliding windows as well as global attention components. \citet{zaheer2020bigbird} proposed \bigbird, another sparse attention mechanism  which combines global attention, random attention and local attention to reduce the complexity to $\bigO(L)$. A similar construction was previously used by \citet{DBLP:journals/corr/abs-2004-08483}. More recently \change{the combination of global and local attention has been utilized by \citet{NEURIPS2021_9425be43} and \citet{NEURIPS2021_f621585d}.  Further}, \citet{xiong2021nystromformer} adapted the Nystr\"{o}m method to approximate standard self-attention with $\bigO(L)$ complexity. \change{\citet{lu2021soft} also makes use of a similar near-field and far-field attention mechanism, where the far field attention is calculated via a low rank approximation using $tanh$ and $ReLU$ non-linearities. A different line of work has attempted to limit the number of key-value pairs that are to be attended to, by attempting to summarise the variable length context with a fixed number of memory-cells~\citep{zhang2022linearizing,NEURIPS2021_14319d9c}. While some of these more nuanced approaches outperform \kerneltransformer, most of their innovations are orthogonal to ours, and} none of these approaches explore kernel learning within the attention mechanism of Transformers. 

\subsection{Kernel Learning}

While kernel methods have long been used to solve non-linear statistical problems, they traditionally scaled poorly with the number of data points thereby limiting their applicability on large datasets~\citep{vapnik1997svm, cortes1995support, scholkopf1998nonlinear, scholkopf2001kernels, hofmann2008kernels}. Prior to \RKS, several kernel approximation techniques have been proposed to improve the scalability of kernel methods, including greedy basis selection techniques~\citep{smola2000greedy}, divide-and-conquer approaches~\citep{hsieh2014divide, zhang2013divide, liu2020learning},  non-stationary spectral kernels \citep{remes2017nonstationary}, generalized spectral kernels \citep{samo2015generalized} as well as Nystr\"{o}m methods~\citep{williams2001nystrom}.

The method of Random Kitchen Sinks has been revisited several times, either to improve the approximation quality~\citep{yu2016orthogonal, chromanski2017orthogonal, li2017normalized, avron2016quasi, lyu2017spherical}, reduce the time and memory complexity \citep{le2013fastfood, choromanski2016recycling, feng2015circluant, dao2017quadrature} or analyze theoretically the risk and generalization properties of the algorithm~\citep{sutherland2015error, sun2018theory, li2019analysis}. A systematic survey of random feature methods for approximating kernel functions can be found in~\citep{liu2021survey}.

Lastly, there exist a class of methods that extend the \RKS\ framework to enable kernel learning. Representative approaches involve either a one-stage~\citep{yang2015alacarte, fang2020endtoend} or a two-stage procedure~\citep{sinha2016learning, li2019implicit, bullins2018notsorandom, shen19harmonizable}. Two-stage approaches involve an intermediate step in which the problem of kernel-alignment is solved~\citep{cristianini2002alignment}. However, solving the kernel-alignment problem requires accessed to labeled data which is not available in this case, as inputs to the kernel learning algorithm are the intermediate representations of the input sequence. 

 \section{Conclusion}
In this paper, we bridged the gap between advances in kernel learning and efficient Transformers by proposing
several kernel learning methods for Transformers that increase the expressiveness of Transformers while keeping the computational complexity linear in sequence length.
We showed that our proposed \kerneltransformer\ are Turing-complete and can control their variance.
Experimentally our proposed models perform on par with, and possibly exceed the performance of existing efficient transformer architectures on long context tasks without falling behind on short context tasks. We also found that for some datasets such as \listops, \RKS\ based models tend to fall short of their PRF counterparts. Our experiments further demonstrate that the memory consumption of our models scales linearly with the sequence length.

\section*{Ethical Considerations}
Our work is on making Transformers computationally efficient without losing expressiveness.
Our models were evaluated on publicly available benchmark datasets. The datasets used in our work do not contain sensitive information to the best of our knowledge. 


\noindent{\bf Reproducibility:}
We plan to open source the entire code of the \kerneltransformer\ framework (including the implementation of all models as well as the code for replicating all of our experiments) before the camera ready version of the paper. As part of this submission, we include code for all the methods proposed by us along with instructions on how to reproduce results.
A detailed description of all hyperparameters (for both \lra\ as well as \glue\ benchmarks) has been included in Appendix~\ref{sec:appen:expt}. Finally, regarding our theoretical contributions, we present a detailed theoretical analysis in Appendix~\ref{sec:Proof}.
\bibliography{tmlr,custom}
\bibliographystyle{tmlr}
\newpage

{
\begin{center}
    \Large
    \textbf{{Learning The Transformer Kernel -- Appendix}}
\end{center}
}
\appendix
\section{Detailed proof of Theorems}
\label{sec:Proof} 
\subsection{Definitions}
\label{Definitions}
\textbf{Transformer:}A transformer consists of an Encoder and a Decoder, which in turn consist of several encoder and decoder layers respectively. A single encoder layer consists of an attention layer($Att$) and a 2 layer feed-forward neural network($O$) :
\begin{align}
\mathbf{a}_i&= Att(W_q\mathbf{x}_i,W_k\mathbf{X},W_v\mathbf{X})+\mathbf{x}_i\\
\mathbf{z}_i&=O(\mathbf{a}_i)+\mathbf{a}_i
\end{align}
In  our case, the feed-forward neural network uses perceptron activations(ie $f_{perc}(x)=1$ iff $x>0$ and $0$ otherwise) and the attention is gaussian(discussed in detail later). The final layer of the encoder is followed by a couple of two layer output neural networks, which produce the Encoder Key($\mathbf{K_e}$) and Encoder Value($\mathbf{V_e}$) to be used by the decoder. In our proof, we assume these to have ReLU activation($f_{ReLU}(x)=max(0,x)$)

The decoder layers are similar to the encoder layer except for an additional cross attention layer which attends to the encoder output:
\begin{align}
\mathbf{p}_i&= Att(W_q\mathbf{y}_i,W_k\mathbf{Y},W_v\mathbf{Y})+\mathbf{y}_i\label{self-att}\\
\mathbf{a}_i&= Att(W_q'\mathbf{p}_i,\mathbf{K_e},\mathbf{V_e})+\mathbf{y}_i\\
\mathbf{z}_i&=O(\mathbf{a}_i)+\mathbf{a}_i
\end{align}
Unlike the encoder, the decoder self attention if Eq. \ref{self-att} can only attend to previous position. After the final layer we have a two layer feed-forward neural network with ReLU activation to produce the output. The decoder is initialised with a special \textit{seed vector} and is repeatedly applied with the right shifted output of the last application as the input of the current application, until some termination condition is fulfilled.

Both the encoder and decoder can further use \textit{position embeddings}, which have the same dimension as the output of each layer, and are added to the input prior to the first layer. These help in establishing the order of the input  

Since the output of any unit of a layer is independent of values to its right, these do not change with time and can be cached. The output of the final layer of the rightmost cell can therefore be regarded as the model state encoding($v$)\\

\noindent\textbf{Turing Machine:}A Turing Machine is an abstract construct which consists of a right infinite tape and a read-write head. Each cell of the tape can hold one of many symbols from a predefined alphabet $\Sigma$ which includes a special blank symbol $b$. Additionally, the read-write head can be in one of many possible states within the state-space $Q$ which includes a special initial state $q_{init}$ and a subset of final states $F$. 

Initially, the tape contains the input followed by an infinite number of blank symbols, while the head starts off in the last non-blank cell. In each step, the head executes in accordance with a transition function $T(s^{(i)},q^{(i)})=(v^{(i)},q^{(i+1)},m^{(i)})$ , ie, based on the symbol currently under the head and the current state, it decides the symbol it wants to overwrite the current symbol with, the state it will be in the next step and the direction it wants to move, which can be either left($-1$) or right($1$). We assume that the transition function already makes sure that the head never moves left from the leftmost cell.

For the purpose of our proof, we additionally define $c^{(i)}$ as the index of the cell to which the head currently points, $\ell(i)$, which represents the step number when the head last pointed to the current cell, ie $\ell(i)=\max\{j|c^{(j)}=c^{(i)}\}$. In the special case where the current cell is being visited for the first time, we have $\ell(i)=i-1$ 
\subsection{The Proof}
In this section, we provide a general proof which works for both \textbf{Theorem 1} and \textbf{Theorem 2} in the paper. This is possible since the construction only makes use of the dual of the kernel functions used, i.e. the gaussian. The fact that both kernel functions map to the gaussian is shown in lemma S.2 (Sec \ref{S.2})\\

Our proof is based on the similar proof in \cite{DBLP:journals/corr/abs-1901-03429}. Any symbols not explicitly defined have same meanings from that paper. We begin the proof by defining our model encoding($\mathbf{v})$:
\begin{align}
    \begin{split}
        \mathbf{v}=[&\mathbf{q}_1,\mathbf{s}_1,x_1,x_2\\
        &\mathbf{q}_2,\mathbf{s}_2, x_3,x_4,x_5,x_6,\\
        &x_7,\mathbf{s}_4,x_8,\\
        &x_{9},x_{10},x_{11},x_{12}]
    \end{split}
\end{align}
where $q_i\in \mathbb{Q}^{|Q|}$, $s_i\in \mathbb{Q}^{|\Sigma|}$, and $x_i\in \mathbb{Q}$, giving a total model size of $2|Q|+3|\Sigma|+14$. Hereafter, $[\![x]\!]$ represents the one-hot encoding for the state $x$ or symbol $x$ depending on the position it is being used in. $\mathbf{0_q}$ represents all $0$'s  in a state field, and represents the $q_{copy}$ state discussed later, while  $\mathbf{0_s}$ represents all $0$'s  in a symbol field, and represents the  blank symbol. Further, $\beta^{(i)}=\min(i,n)$  where $n$ is the size of the encoder and $\alpha^{(i)}$ represents the symbol at position $\beta^{(i)}$ in the encoder. We assume that atleast the last cell of the encoder contains a blank symbol.

This differs from \cite{DBLP:journals/corr/abs-1901-03429} in the addition of a fourth scalar in  the first group, in which we intend to store the current position $c^{(i)}$ of the head. \\
Our invariant  is that $\mathbf{y}_i$ the output from the decoder at timestep $i$, stores:
\begin{enumerate}
    \item The current state of the Turing Machine($q^{(i)}$)
    \item The symbol under the head($s^{(i)}$)
    \item The direction of movement of the head in the previous timestep ($m^{(i-1)}$)
    \item The current position of the head($c^{(i)}$)
\end{enumerate}
In all, we get $\mathbf{y}_i=[[\![q^{(i)}]\!],[\![s^{(i)}]\!],m^{(i-1)},c^{(i)},0,\ldots,0] $\\\\
\textbf{Positional Embeddings:} The last group ($x_9,x_{10},x_{11},x_{12}$) is dedicated to the positional embeddings, which are given as($1,i,\frac{1}{i},\frac{1}{i^2}$) These same embeddings are added on both the Encoder and Decoder side.\\\\
\textbf{Encoder}: The encoder consists of a single layer. It gets as input the symbol at position $i$ and the positional embeddings, ie $input_i=[\mathbf{0}_q,\mathbf{0}_s,0,0,\mathbf{0}_q,,[\![s^{(i)}]\!],0,0,0,0,i,\mathbf{0}_s,0,1,i,\frac{1}{i},\frac{1}{i^2}]$ which has a trivial attention layer(ie, one that outputs all zeroes) and a feed forward layer which separates the positional embeddings from the symbols, giving $\mathbf{k}^e_i=[0,\ldots,0,i,-1,0,0]$ and $\mathbf{v}^e_i=[\mathbf{0}_q,\mathbf{0}_s,0,0,\mathbf{0}_q,,[\![s^{(i)}]\!],0,0,0,0,i,\mathbf{0}_s,0,0,0,0,0]$.\\\\ 
\textbf{Decoder Layer 1}: The first layer of the decoder calculates the next state, the symbol to be written and the direction of movement of the head. This includes 2 cases: 
\begin{enumerate}
    \item Initially, the Decoder starts off with in the state $q_{copy}$. While the state is still $q_{copy}$, the head writes the symbol at the $i^{th}$ position in  the encoder and moves right, until a blank symbol is seen. Once a blank symbol is reached, the tape rewrites the blank symbol, moves left and the state changes to $q_{init}$.
    \item Once we move into $q_{init}$, the output is fully defined by the current state and symbol under the head.
\end{enumerate}
To facilitate the first case, we make use of the cross attention layer, to get 

\begin{align}\begin{split}
    Att(\mathbf{q},\mathbf{K^e},\mathbf{V^e})&=[0,\ldots,0,\\
    &\hspace{1.6em} \mathbf{0}_q, [\![\alpha^{(i)}]\!], 0, 0, 0 ,0 ,\\
    &\hspace{1.6em} \beta^{(i)},\mathbf{0_s},0,\\
    &\hspace{1.6em} 0,0,0,0]\\
    &=\mathbf{v}^e_{\beta^{(i)}}
\end{split}\end{align}
The details of this process are explained in lemma S.1(see sec. \ref{S.1})  Adding in the residual connection, we have:
\begin{align}\begin{split}
    \mathbf{a_i^1}&=[[\![q^{(i)}]\!],[\![s^{(i)}]\!],m^{(i-1)},c^{(i)}\\
    &\hspace{1.6em} \mathbf{0}_q, [\![\alpha^{(i)}]\!], 0, 0, 0 ,0 ,\\
    &\hspace{1.6em} \beta^{(i)},\mathbf{0_s},0,\\
    &\hspace{1.6em}1,i+1,\frac{1}{(i+1)},\frac{1}{(i+1)^2}]
\end{split}\end{align}
Hereafter, we make use of the feed-forward layer to get:
\begin{align}
    \begin{split}
        O(\mathbf{a_i^1})&=[-[\![q^{(i)}]\!],-[\![s^{(i)}]\!],-m^{(i-1)},m^{(i)},\\
    &\hspace{1.6em} [\![q^{(i+1)}]\!],[\![\bar{v}^{(i)}]\!],m^{(i)},m^{(i-1)},0, 0,\\
    &\hspace{1.6em} 0,\ldots,0\\
    &\hspace{1.6em} 0,\ldots,0]
    \end{split}
\end{align}
If the state is $q_{init}$ then we set $[\![\bar{v}^{(i)}]\!]=\mathbf{0}_s$, else we have $[\![\bar{v}^{(i)}]\!]=[\![v^{(i)}]\!]$. Note that this gives us  $[\![\bar{v}^{(i)}]\!]+ [\![\alpha^{(i)}]\!]=[\![v^{(i)}]\!]$. 

To get all the required values, we first project$[\![q^{(i)}]\!]$ and $[\![s^{(i)}]\!]$ to a one-hot encoding of $Q\times\Sigma$. from there, we can calculate all the required values in a look-up table fashion.  if the state is $q_{init}$ then we set $[\![v^{(i)}]\!]=\mathbf{0}_s$   \\\\The final output of this layer is then:
\begin{align}
    \begin{split}
        \mathbf{z_i^1}&=[0,\ldots,0,c^{(i+1)}\\
    &\hspace{1.6em} [\![q^{(i+1)}]\!],[\![v^{(i)}]\!],m^{(i)},m^{(i-1)},0,0,\\
    &\hspace{1.6em}\beta^{(i)},\mathbf{0_s},0,\\
    &\hspace{1.6em}1,i+1
    ,\frac{1}{(i+1)},\frac{1}{(i+1)^2}]
    \end{split}
\end{align}

\textbf{Decoder Layer 2:} In this layer we calculate the symbol under the head in the next timestep. In order to do so, we first use the self attention layer to calculate $[\![v^{(\ell(i+1))}]\!]$ and $(\ell(i+1))$(For details, see sec \ref{S.3}.):
\begin{align}\begin{split}
    Att(W^2_q\mathbf{z}^2_i,W^2_k\mathbf{Z}^2,W^2_v\mathbf{Z}^2)&=[0,\ldots,0,\\
    &\hspace{1.6em} 0,\ldots,0,\\
    &\hspace{1.6em} 0,[\![v^{(\ell(i+1))}]\!],(\ell(i+1)),\\
    &\hspace{1.6em} 0,0,0,0]
\end{split}\end{align}
Adding the residual layer, we have
\begin{align}
    \begin{split}
        \mathbf{a_i^2}&=[0,\ldots,0,c^{(i+1)}\\
    &\hspace{1.6em} [\![q^{(i+1)}]\!],[\![v^{(i)}]\!],m^{(i)},m^{(i-1)},0,0,\\
    &\hspace{1.6em} \beta^{(i)},[\![v^{(\ell(i+1))}]\!],(\ell(i+1)),\\
    &\hspace{1.6em}1,i+1,\frac{1}{(i+1)},\frac{1}{(i+1)^2}]
    \end{split}
\end{align}
The feed-forward layer then gives $O_2(\mathbf{a_i^2})=[ [\![q^{(i+1)}]\!],[\![v^{(\ell(i+1))}]\!]-f_{perc}((\ell(i+1)+2-(i+1)),m^{(i-1)},0,-M,\ldots-M] $ where  M is a large negative value. The perceptron function in the $\mathbf{s}_1$ is added positionwise, and is $0$ unless $\ell(i+1)=i$. In this special case, it makes  $\mathbf{s}_1$ contain only $0$ or $-1$ which is converted into $\mathbf{0}_s$ by the ReLU activation in the output MLP. The same is also true for every field after the first $4$, where we add a large negative value to make the ReLU output $0$.
\subsection{The Attention Mechanism}
The attention mechanism in the Gaussian kernel is defined as follows:
\begin{align}
    Attn(\mathbf{Q},\mathbf{K},\mathbf{V})&=\mathbf{V}\big(ColNorm\big(\Phi(\mathbf{Q})^T\Phi(\mathbf{K})\big)\big)\label{att}
\end{align}
where $\Phi$ is $\Phi_{RKS}$ (Eq \ref{eq:RKSAG}) for Theorem 1 and  $\Phi_{PRF}$ (Eq \ref{eq:PRFAG}) for Theorem 2, and $\omega_i$ is sampled from a gaussian with zero mean and diagonal covariance. However, for the proof construction, we use a hard version of this attention mechanism, and limit ourselves to the standard gaussian for $\omega$(since the mean and sigma is learnable, this can always be achieved). To begin with, we replace the kernels with their common dual, using lemma S.2 (sec \ref{S.2}). In our construction, we do not require learnable means and variances, so we fix them  to be $0_{d_q}$ and $\mathbb{I}$ hereafter:
\begin{align}
Attn(\mathbf{Q},\mathbf{K},\mathbf{V})&=\sum_{i=0}^{h-1}\mathbf{W}_O^i\mathbf{V}\Big(ColNorm\Big[\!\Big[e^{-\frac{||\mathbf{q}_l-\mathbf{k}_m||^2}{2}}\Big]\!\Big]_{l=0,m=0}^{d-1,d-1}\Big)
\end{align}
where $[\![f(l,m)]\!]_{l=0,m=0}^{\alpha,\beta}$ denotes an $\alpha \times\beta$  matrix whose $(l,m)^{th}$ entry is $f(l,m)$, $ColNorm(\mathbf{X})$ indicates the matrix $\mathbf{X}$ with its columns normalised to  and $d$ is the dimension of the query/key vector.

While this definition does not seem to allow multiplying the exponent, one must remember that the query and key matrices are calculated using projection matrices, and any required scalar factor can be incorporated into them. Therefore, we define hard gaussian attention as:
\begin{align}
    \label{Att}
    score(\mathbf{u},\mathbf{v})=-||\mathbf{u}-\mathbf{v}||^2
\end{align}
Hard attention is them computed as 
\begin{align}
    \label{Att1}
    Att(\mathbf{q_i},\mathbf{K},\mathbf{V})=\frac{\sum_{j=0}^{n-1}\mathbb{I}[score(\mathbf{q_i},\mathbf{k_j})=(max_{j'}score(\mathbf{q_i},\mathbf{k_{j'}}))]\mathbf{v_j}}{\sum_{j=0}^{n-1}\mathbb{I}[score(\mathbf{q_i},\mathbf{k_j})=max_{j'}(score(\mathbf{q_i},\mathbf{k_{j'}}))]}
\end{align}
Here $\mathbb{I}$ in  the indicator function. 
\subsection{Lemma S.1}
\label{S.1}
\subsection{Statement}
Given
\begin{align}
\begin{split}
    \mathbf{q}&=[\_\_,\ldots,\_\_,1,i,\_\_,\_\_]\\
    \mathbf{k}^e_j&=[0,\ldots,0,\\
    &\hspace{1.6em} 0,\ldots,0,\\
    &\hspace{1.6em} 0,\ldots 0,\\
    &\hspace{1.6em} j,-1,0,0]\\
    \mathbf{v}^e_j&=[0,\ldots,0,\\
    &\hspace{1.6em} \mathbf{0}_q, [\![s^{(j)}]\!], 0, 0, 0 ,0 ,\\
    &\hspace{1.6em} j,\mathbf{0_s},0,\\
    &\hspace{1.6em} 0,0,0,0]\\
\end{split}\end{align}
    For $j\in \{0,\ldots,n\}$, we need a construction that gives
\begin{align}\begin{split}
    Att(\mathbf{q},\mathbf{K^e},\mathbf{V^e})&=[0,\ldots,0,\\
    &\hspace{1.6em} \mathbf{0}_q, [\![\alpha^{(i)}]\!], 0, 0, 0 ,0 ,\\
    &\hspace{1.6em} \beta^{(i)},\mathbf{0_s},0,\\
    &\hspace{1.6em} 0,0,0,0]\\
    &=\mathbf{v}^e_{\beta^{(j)}}
\end{split}\end{align}
\subsubsection{Proof}
Note that while the key and value comes from the encoder, and is therefore fixed, the query comes from the decoder and thus can be projected  as we please. It is easy to construct a projection  matrix that gives $W_Q\mathbf{q}=[0,\ldots,i,-1,0,0]$. Then we have $score(\mathbf{q},\mathbf{k}_{j'})=-||i=j||^2=-(i-j)^2$, whose  maxima on $j'$ is unique and occurs at $i=\beta^{(j)}$. Thus, we have $Att(\mathbf{q},\mathbf{K^e},\mathbf{V^e})=\mathbf{v}^e_{\beta^{(j)}}$, which is exactly what we wanted.
\subsection{Lemma S.2}
\label{S.2}
\subsubsection{Statement}
Let
\begin{align}
  \Phi_{PRF}(\mathbf{x})&=[\exp(-||\mathbf{x}||+\omega_0^T\mathbf{x}),\ldots,\exp(-||\mathbf{x}||+\omega_{k-1}^T\mathbf{x})]\label{eq:PRFAG}\\
  \Phi_{RKS}(\mathbf{x})&= \sqrt{\frac{2^m}{k}}[\cos(\omega_1^Tx),\ldots,\cos(\omega_k^Tx), \sin(\omega_1^Tx),\ldots,\sin(\omega_k^Tx)]\label{eq:RKSAG}
\end{align} 

We want to show that if $\omega\sim \mathcal{N}(0,\mathbf{I})$ then the kernels as defined above corresponds to the \textit{GMM kernel}, ie. 
\begin{align}
    \Phi_{RKS}({\mathbf{X}})\Phi_{RKS}({\mathbf{Y}})\approx\Phi_{PRF}({\mathbf{X}})\Phi_{PRF}({\mathbf{Y}})\approx e^{-\frac{||\mathbf{x-y}||^2}{2}}
\end{align}
\subsubsection{Proof}

\textbf{Positive Random Features:}The proof actually follows from a trivial extension of Lemma 1 in \citep{choromanski2021performers}, but we present it here end to end for the convenience of the reader.\\

First we observe that
\begin{align}\begin{split}
    e^{-\frac{||\mathbf{x-y}||^2}{2}}&=e^{-\frac{||\mathbf{x}||^2-2\mathbf{x^Ty}+||\mathbf{y}||^2}{2}}\\
    &=e^{-\frac{2*||\mathbf{x}||^2-||\mathbf{x}||^2-2\mathbf{x^Ty}+2*||\mathbf{y}||^2-||\mathbf{y}||^2}{2}}\\
    &=e^{-||\mathbf{x}||^2}e^{\frac{||\mathbf{x+y}||^2}{2}}e^{-||\mathbf{y}||^2}
    \label{eq:der1}
\end{split}\end{align}

Next we leverage the fact that $(2\pi)^{-d_q/2}\int e^{-\frac{||\omega-\mathbf{c}||^2}{2}}d\omega=1$ in to evaluate the second factor above:
\begin{align}
    \begin{split}
        e^{\frac{||\mathbf{x+y}||^2}{2}}
        &=(2\pi)^{-d_q/2}e^{\frac{||\mathbf{x+y}||^2}{2}}\int e^{-\frac{||\omega-\mathbf{x+y}||^2}{2}}d\omega\\
        &=(2\pi)^{-d_q/2}\int e^{-\frac{||\omega||^2+||\mathbf{x+y}||^2-||\mathbf{x+y}||^2-2\omega^T\mathbf{x}-2\omega^T\mathbf{y}}{2}}d\omega\\
        &=(2\pi)^{-d_q/2}\int e^{-\frac{||\omega||^2}{2}}e^{\omega^T\mathbf{x}}e^{\omega^T\mathbf{y}}d\omega\\
        &=\mathbb{E}_{\omega_i\sim\mathcal{N}(0,\mathbf{I})}(e^{\omega^T\mathbf{x}}e^{\omega^T\mathbf{y}})
    \end{split}
\end{align}

The terms $e^{-||\mathbf{x}||^2}$ and $e^{-||\mathbf{y}||^2}$ in the last line of Eq.\ref{eq:der1} are independent of $\omega$ and can thus be pushed into the expectation. Finally, we approximate the expectation by sampling in order to get the required result.\\
\textbf{Random Kitchen Sinks:}For the second kernel, we start with Eq. 7.4.6 in  \cite{abramowitz1972handbook} and extend it to vectors.  We have,

\begin{align*}\begin{split}
    &\quad\int_{\mathbb{R}^m}e^{-||t||^2}\cos(2\mathbf{t}^T\mathbf{x})d\mathbf{t}\\
    &=\int_{\mathbb{R}^m}e^{-(t_0^2+\sum_{i=1}^{m-1}t_i^2)}\cos\Big(2x_0t_0+2\sum_{i=1}^{m-1}t_ix_i\Big)dt_0dt_1\ldots dt_{m-1}\\
    &=\int_{\mathbb{R}^m}e^{-(t_0^2+\sum_{i=1}^{m-1}t_i^2)}\cos(2x_0t_0)\cos\Big(2\sum_{i=1}^{m-1}t_ix_i\Big)dt_0dt_1\ldots dt_{m-1}\\
    &\quad-\int_{\mathbb{R}^m}e^{-(t_0^2+\sum_{i=1}^{m-1}t_i^2)}\sin(2x_0t_0)\sin\Big(2\sum_{i=1}^{m-1}t_ix_i\Big)dt_0dt_1\ldots dt_{m-1}
\end{split}\end{align*}
\noindent The second integral involving $\sin$  is odd and therefore evaluates to 0. That leaves us with:
\begin{align*}\begin{split}
&=\int_{\mathbb{R}^m}e^{-(t_0^2+\sum_{i=1}^{m-1}t_i^2)}\cos(2x_0t_0)\cos\Big(2\sum_{i=1}^{m-1}t_ix_i\Big)dt_0dt_1\ldots dt_{m-1}\\
&=\int_{\mathbb{R}^{m-1}}\bigg(\int_{-\infty}^\infty e^{-t_0^2}\cos(2x_0t_0)dt_0\bigg)e^{-\sum_{i=1}^{m-1}t_i^2}\cos\Big(2\sum_{i=1}^{m-1}t_ix_i\Big)dt_1\ldots dt_{m-1}\\
&=\frac{1}{2}\sqrt{\pi}e^{-x_0^2}\int_{\mathbb{R}^{m-1}}e^{-\sum_{i=1}^{m-1}t_i^2}\cos\Big(2\sum_{i=1}^{m-1}t_ix_i\Big)dt_1\ldots dt_{m-1}\\
\end{split}\end{align*}
This process can now be repeated for every dimension of $t$ and $x$ to finally give:
\begin{align}
    \int_{\mathbb{R}^m}e^{-||\mathbf{t}||^2}\cos(2\mathbf{t}^T\mathbf{x})d\mathbf{t}=\frac{\pi^{m/2}}{2^m}e^{-||\mathbf{x}||^2}\label{eq:asvec}
\end{align}

\noindent Using Eq. \ref{eq:asvec}, we can now get
\begin{align}
    \begin{split}
        \Phi_{RKS}({\mathbf{X}})\Phi_{RKS}({\mathbf{Y}})&=\frac{2^{m}}{k}\sum_{i=0}^{k-1}\big(\cos(\mathbf{\omega}_i^T\mathbf{x})\cos({\omega}_i^T\mathbf{y})+\sin(\mathbf{\omega}_i^T\mathbf{x})\sin({\omega}_i^T\mathbf{y})\big)\\
        &=\frac{2^{m}}{k}\sum_{i=0}^{k-1}\cos\big(\mathbf{\omega}_i(\mathbf{x}-\mathbf{y})\big)\\
        &\approx2^{m}\mathop{\mathbb{E}}_{\mathbf{\omega}_i\sim \mathcal{N}(\mathbf{0}_m,\mathbb{I})}\cos\big(\mathbf{\omega}(\mathbf{x}-\mathbf{y})\big)
        \\
        &=2^{m}\int_{\mathbb{R}^m}\frac{1}{(2\pi)^{m/2}}e^{-\frac{||\mathbf{\omega}||^2}{2}}\cos(2\mathbf{\omega}^T\frac{\mathbf{x-y}}{2})d\mathbf{\omega}\\
         &=\frac{2^{m}}{\pi^{m/2}}\int_{\mathbb{R}^m}e^{-||\mathbf{\frac{\omega}{\sqrt{2}}||^2}}\cos(2\mathbf{\frac{\omega}{\sqrt{2}}}^T\frac{\mathbf{x-y}}{\sqrt{2}})d\mathbf{\frac{\omega}{\sqrt{2}}}\\
        &=e^{-\frac{||\mathbf{x-y}||^2}{2}}
    \end{split}
\end{align}

In either case, the error stems from the approximation of the expectation by sampling, which can be made arbitrarily small by increasing $k$

\subsection{Lemma S.3}
\label{S.3}
\subsubsection{Statement}
Given,
\begin{align}
\begin{split}
    \mathbf{z}^1_i&=[0,\ldots,0,c^{(i+1)},\\
    &\hspace{1.6em} [\![q^{(i+1)}]\!],[\![v^{(i)}]\!],m^{(i)},m^{(i-1)},0,0,\\
    &\hspace{1.6em} \beta^{(i+1)},\mathbf{0_s},0,\\
    &\hspace{1.6em}1,(i+1),\frac{1}{(i+1)},\frac{1}{(i+1)^2}]
\end{split}
\end{align}
we need a construction  that gives 
\begin{align}\begin{split}
    Att(W^2_q\mathbf{z}^2_i,W^2_k\mathbf{Z}^2,W^2_v\mathbf{Z}^2)&=[0,\ldots,0,\\
    &\hspace{1.6em} 0,\ldots,0,\\
    &\hspace{1.6em} 0,[\![v^{(\ell(i+1))}]\!],(\ell(i+1)),\\
    &\hspace{1.6em} 0,0,0,0]
\end{split}\end{align}
\subsubsection{Proof}
We set the weight matrices to get $\mathbf{q}_j=W^2_q\mathbf{z}^2_j=[0,\ldots,0,c^{(j+1)},0,0]$, $\mathbf{k}_j=W^2_k\mathbf{z}^2_j=[0,\ldots,0,c^{(j)}=c^{(j+1)}-m^{(i)},0,\frac{1}{(j+1)}]$ and $\mathbf{v}_j=W^2_v\mathbf{z}^2_j=[0,\ldots,0,[\![v^{(j)}]\!],j,0,0,0,0]$. All these are partial permutations and therefore can be done using appropriate binary matrices.\\

Note that the required output is exactly the value at $j=\ell(i+1)$, so it is sufficient to show that the $score(\mathbf{q}_i,\mathbf{k}_j)$ is maximised in $j$ for $j=\ell(i+1)$, i.e. 
\begin{align}
j=
\begin{cases}
\max\{j'|c^{(j')}=c^{(i+1)}\}, &if\;\;\exists j'\; s.t.\; c^{(j')}=c^{(i+1)}\\
i, &\text{otherwise}
\end{cases}
\end{align}
Now  we have $score(\mathbf{q}_i,\mathbf{k}_j)=-(c^{(i+1)}-c^{(j)})^2-\frac{1}{(j+1)^2}$. For all $j$ such that $c^{(i+1)}\neq c^{(j)})$, the $score$ is almost $-1$ since $c$ is an integer. If there $\exists j'\; s.t.\; c^{(j')}=c^{(i+1)}$ then the corresponding $score$ is greater that $-1$, and the maxima is achieved at the highest such value of $j$. If such $j$ does not exist however, then $\forall j<i,\;score(\mathbf{q}_i,\mathbf{k}_j)<-1-\frac{1}{(i+1)^2}$ and therefore, the maxima is achieved at $j=i$.

\section{Mean Square Error of Linear Approximations}
\label{sec:MSE}
In this section, we calculate the Variance/Mean Square Error(MSE) in the Linear approximation of the Gaussian Kernel. Our proof is based on the similar proof in \citet{choromanski2021performers}.
\subsection{Random Kitchen Sinks}
\subsubsection{Statement}
For a \MG\ estimator with $m$ samples for a Normal distribution with mean vector $\mu$ and Covariance Matrix $\Sigma=S^TS$, the variance of the estimate around its mean is given by:
\begin{equation}
    MSE(\phi_{\MG}(q)^T\phi_{\MG}(k))=\frac{2}{m}\cos^2(\mu^T(k-q))(1-\exp(-||
S^T(k-q)||^2))^2
\end{equation}
\subsubsection{Proof}
\begin{align}
    \begin{split}
        &\quad MSE(\phi_{\MG}(q)^T\phi_{\MG}(k))\\&=\frac{1}{m^2}Var_{\omega_i\sim \mathcal{N}(\mu,\Sigma),\chi_i\sim \mathcal{N}(-\mu,\Sigma)}(\sum_{i=1}^m(\cos(\omega_i^Tq)\cos(\omega_i^Tk)+\sin(\omega_i^Tq)\sin(\omega_i^Tk)\\&\qquad\qquad\qquad\qquad\qquad\qquad\qquad+\cos(\chi_i^Tq)\cos(\chi_i^Tk)+\sin(\chi_i^Tq)\sin(\chi_i^Tk)))\\
        &=\frac{1}{m^2}Var_{\omega_i\sim \mathcal{N}(\mu,\Sigma),\chi_i\sim \mathcal{N}(-\mu,\Sigma)}(\sum_{i=1}^m(\cos(\omega_i^T(q-k))+\cos(\chi_i^T(q-k)))\\
        &=\frac{1}{m^2}Var_{\eta_i\sim\mathcal{N}(0,\mathbb{I})}(\sum_{i=1}^m(\cos((\eta_i^TS^T+\mu^T)(q-k))+\cos((\eta_i^TS^T-\mu^T)(q-k))))\\
        &=\frac{4}{m^2}Var_{\eta_i\sim\mathcal{N}(0,\mathbb{I})}(\sum_{i=1}^m\cos(\eta_i^TS^T(q-k))\cos(\mu^T(q-k)))\\
        &=\frac{4}{m^2}cos^2(\mu^T(q-k))Var_{\eta_i\sim\mathcal{N}(0,\mathbb{I})}(\sum_{i=1}^m\cos(\eta_i^TS^T(q-k)))\\
        &=\frac{4}{m}cos^2(\mu^T(q-k))Var_{\eta\sim\mathcal{N}(0,\mathbb{I})}\cos(\eta^T(S^T(q-k)))\\
        &=\frac{2}{m}cos^2(\mu^T(q-k))(1-\exp(-||S^T(q-k)||^2)^2)
    \end{split}
\end{align}
Note that there is a slight abuse of notation in that $\omega$ and $\chi$ are not independently sampled, but are transforms of the same sample, making the change of variable valid. Since all $\eta$ are iid, we can pull the summation out of the variance. Thereafter we apply Lemma 1 from \citet{NIPS2016_53adaf49}  to calculate the final variance.
\subsection{Positive Random Features}
For a \PRF\ estimator with $m$ samples for a Normal distribution with mean vector $\mu$ and Covariance Matrix $\Sigma=S^TS$, the variance of the estimate around its mean is given by:

\begin{align}
    \begin{split}
     &\quad MSE(\phi_{\PRF}(q)^T\phi_{\PRF}(k))\\&=\frac{1}{m}\exp(-2(||q||^2+||k||^2-\mu^T(q+k)))(\exp(2||S^T(q+k)||)-\exp(||S^T(q+k)||))
    \end{split}
\end{align}
\subsubsection{Proof}
\begin{align}
    \begin{split}
        &\quad MSE(\phi_{\PRF}(q)^T\phi_{\PRF}(k))\\&=\frac{1}{m^2}Var_{\omega_i\sim \mathcal{N}(\mu,\Sigma)}(\sum_{i=1}^m(\exp(\omega_i^T(q+k)-||q||^2-||k||^2)))\\
        &=\frac{1}{m}\exp(-2(||q||^2+||k||^2))Var_{\omega\sim \mathcal{N}(\mu,\Sigma)}\exp(\omega^T(q+k))\\
        &=\frac{1}{m}\exp(-2(||q||^2+||k||^2))Var_{\eta\sim \mathcal{N}(0,\mathbb{I})}(\exp(\eta^TS^T(q+k)+\mu^T(q+k)))\\
        &=\frac{1}{m}\exp(-2(||q||^2+||k||^2-\mu^T(q+k))).\\
        &\qquad\quad(\mathbb{E}_{\eta\sim \mathcal{N}(0,\mathbb{I})}(\exp(2\eta^TS^T(q+k)))-(\mathbb{E}_{\eta\sim \mathcal{N}(0,\mathbb{I})}^2(\exp(\eta^TS^T(q+k)))))\\
        &=\frac{1}{m}\exp(-2(||q||^2+||k||^2-\mu^T(q+k)))(\exp(2||S^T(q+k)||)-\exp(||S^T(q+k)||))
    \end{split}
\end{align}
Where the last step follows from Eq. $16$ in \citet{choromanski2021performers} which in turn follows from the fact that \PRF\ is an unbiased estimator for gaussians. 
\clearpage

\section{Experimental Details\label{sec:appen:expt}}

\subsection{Source Code}

We implemented \kerneltransformer s in Python 3 and PyTorch \citep{paszke2019torch} and plan to open-source the code for reproducing all experiments upon acceptance.

\subsection{LRA image dataset results}
For completeness we also give results on \image\ datasets from LRA task where a $N \times N$ image is flattened into a sequence of $N^2$ pixels which is then provided as input to the model. The gray-scaled CIFAR10 image classification dataset \citep{krizhevsky09images} is used, resulting in a sequence length of $1024$.

\begin{table}[!t]
\begin{center}
  \begin{tabular}{@{}l|r@{}}
    \toprule
    \textbf{Model} & \image \\
    \midrule\midrule

    Random Predictor  & 10.00  \\
    \midrule
    \multicolumn{2}{c}{Baseline Models}  \\
    \midrule
    \vanillatrans (\citeauthor{vaswani2017transformers}) & 42.44 \\
    \synthesizer (\citeauthor{tay2021synthesizer}) & 41.61 \\
    \sinkhorn (\citeauthor{tay2020sinkhorn}) & 41.23 \\
    Sparse Trans.(\citeauthor{child2019sparse}) & \textbf{44.24} \\
    \reformer (\citeauthor{kitaev2020reformer}) & 38.07\\
    Local Attention (\citeauthor{parmar2018image}) & 41.46 \\
    \longformer (\citeauthor{beltagy2020longformer}) & 42.22\\
    \linformer (\citeauthor{wang2020linformer}) & 38.56\\
    \bigbird (\citeauthor{zaheer2020bigbird}) & 40.83\\
    \LE (\citeauthor{katharopoulos2020linear})  & 42.34\\
    \PF (\citeauthor{choromanski2021performers}) & \underline{42.77} \\
    \midrule
    \multicolumn{2}{c}{Kernelized Transformers}  \\
    \midrule
    \MG\ (Eq. \ref{eq:self-attention-mixgauss-linear}) & 42.33 \\
    \FRKS\  (Eq. \ref{eq:fastfood}) & 36.74  \\
    \GEN\ (Eq. \ref{eq:generative-rks})     & 39.84 \\
    \PRF\  (Eqs. \ref{eq:prf-def}, \ref{eq:fgmm} )  & 39.94 \\
    \FPRF\   (Eqs. \ref{eq:prf-def}, \ref{eq:fff} ) & 38.31 \\
    \GenP\ (Eqs. \ref{eq:prf-def}, \ref{eq:fgen} ) & 40.01\\
    \bottomrule
  \end{tabular}
  \caption{Experimental results on the \image\ dataset with 1024 tokens from \lra\ benchmark.}
  \label{table:lra-results1}
  \end{center}
\end{table}

\subsection{Hyperparameters for LRA Tasks}

\begin{table}[!h]
  \centering
  \begin{tabular}{l|cccc}
    \toprule
    \textbf{Parameter} & \textbf{ListOps} & \textbf{Text} & \textbf{Retrieval} & \textbf{Image} \\
    \midrule
    Batch Size    & $32$ & $32$ & $32$ & $256$ \\
    Learning Rate & $5\times 10^{-3}$ & $5\times 10^{-2}$ & $5\times 10^{-2}$ & $5\times 10^{-4}$  \\
    Training Steps/Epochs & $10$K/NA & $20$K/NA & $5$K/NA & NA/200 \\
    Optimizer     & \multicolumn{4}{c}{Adam with Weight Decay ($\beta_1 = 0.9 , \beta_2= 0.98$)}\\
    Weight Decay  & $0.1$ & $0.1$ & $0.1$ & $0.0$ \\
    Warmup Steps  & $1000$ & $8000$ & $8000$ & $175$\\
    Scheduler  & Sqrt Decay & Sqrt Decay & Sqrt Decay & Cosine Decay \\
    Loss          & \multicolumn{4}{c}{Cross Entropy}\\
    Sequence Length   & $2000$ & $4000$ & $4000$ & $1024$\\
    Num. Layers   & $6$ & $4$ & $4$ & $1$\\
    Num. Heads    & $8$ & $4$ & $4$ & $8$\\
    Embedding Dim.& $512$ & $256$ & $128$ & $128$ \\
    Key/Query/Value Dim. & $64$ & $64$ & $32$ & $8$ \\
    Feedforward Dim. & $2048$ & $1024$ & $512$ & $128$ \\
    Dropout Rate & $0.1$ & $0.1$ & $0.1$ & $0.3$ \\
    Activation Function  & Gelu & Gelu & Gelu & Gelu \\
    Positional Encoding  & Sinusoidal & Sinusoidal & Sinusoidal & Learnable \\
    Pooling Mode & CLS & CLS & CLS & CLS  \\
    \bottomrule
  \end{tabular}
  \caption{Hyperparameters for \lra\ tasks.}
  \label{table:lra-hyperparameters}
\end{table}

\begin{table}[!h]
  \centering
  \begin{tabular}{l|cccc}
    \toprule
    \textbf{Model} & \textbf{ListOps} & \textbf{Text} & \textbf{Retrieval} & \textbf{Image}   \\
    \midrule
    \MG         & $256$ & $128$ & $64$ & $128$ \\
    \FRKS    & $64$  & $64$  & $32$  & $8$ \\
    \GEN      & $256$ & $256$ & $128$ & $128$\\
    \PRF        & $256$ & $256$ & $128$ & $128$ \\
    \FPRF        & $64$ & $64$ & $32$ & $8$ \\
    \GenP        & $128$ & $128$ & $64$ & $8$ \\
    \bottomrule
  \end{tabular}
  \caption{Number of random samples $M$ used within each \kerneltransformer .}
  \label{table:lra-number-samples}
\end{table}

\paragraph{Further Notes:}
\begin{itemize}
    \item To benchmark memory in Figure \ref{fig:lra-memory-accuracy}, we used a batch size of $32$ for \txt\ and a batch size of $2$ for \retrieval.
    \item For \MG\ and \PRF\ the number of components $C$ in the mixture was set to $C=2$.
\end{itemize}

\clearpage

\subsection{Hyperparameters for GLUE Tasks}

\begin{table}[!h]
  \centering
  \begin{tabular}{l|c}
    \toprule
    \textbf{Parameter} & \textbf{Value(s)}  \\
    \midrule
    Pre-Training Batch Size    & $64$ \\
    Batch Size    & $64$ \\
    Pre-Training Learning Rate ($\eta_{\textmd{pre}}$) & $5\times 10^{-4}$  \\
    Pre-Training Learning Rate at Step $i$ & $\min(\frac{i}{10000},\frac{I_{\textmd{pre}}-i}{I_{\textmd{pre}}-10000})*\eta_{\textmd{pre}}$  \\
    Training Learning Rate ($\eta_{\textmd{train}}$) & $\{2\times 10^{-3}, 1\times 10^{-4}, 5\times 10^{-4}, 2\times 10^{-5}, 5\times 10^{-6}\}$  \\
    Training Learning Rate at Step $i$ ($\eta_{\textmd{train}}$) & $\min(\frac{10i}{I_{\textmd{tune}}},\frac{I_{\textmd{tune}}-i}{0.9*I_{\textmd{tune}}})*\eta_{\textmd{train}}$  \\
    Pre-Training Epochs & 5 \\
    Training Epochs & 10 \\
    Optimizer     & Adam with Weight Decay ($\beta_1 = 0.9 , \beta_2= 0.999$)\\
    Weight Decay  & $0.01$ \\
    Loss          & Cross Entropy\\
    Sequence Length   & $512$ \\
    Num. Layers   & $3$\\
    Num. Heads    & $10$\\
    Embedding Dimension & $300$\\
    Key/Query/Value Dimension & $64$ \\
    Transformer Feedforward Dimension & $512$\\
    Classifier Feedforward Dimension & $128$\\
    Dropout Rate & $0.1$\\
    Transformer Activation Function  & Gelu \\
    Classifier Activation Function  & Tanh \\
    Positional Encoding  & Sinusoidal \\
    Pooling Mode & CLS \\
    Num. of Samples from Distribution &   $\{64,128\}$\\          
    \bottomrule
  \end{tabular}
  \caption{Hyperparameters for \glue\ tasks. Where multiple parameters were tried, they are listed in curly brackets. $I_{\textmd{pre}}$ denotes the total number of pre-training steps, whereas $I_{\textmd{tune}}$ denotes the total number of fine-tuning steps on each \glue\ task.}
  \label{tab:glue-params}
\end{table}
Our model has significantly fewer number of parameters as compared to \citet{devlin2019bert} and therefore we perform poorer on than them on all datasets. They use $24$ layers with $16$ heads each. If reported in the same order as the columns of Table \ref{tab:GLUE},  their numbers would look like: $97.5,\,89.3/85.4,\,72.1/89.3,\,87.2/86.4,\,92.7,\,65.1,\,70.1$. 

While we had to limit model sizes due to resource limitations, this handicaps all models equally, and therefore should not prevent comparison across various models reported in our paper.

\subsection{Further Results on Efficiency Benchmarks}
\begin{figure}[h]
    \centering 

        \includegraphics[width=0.47\textwidth]{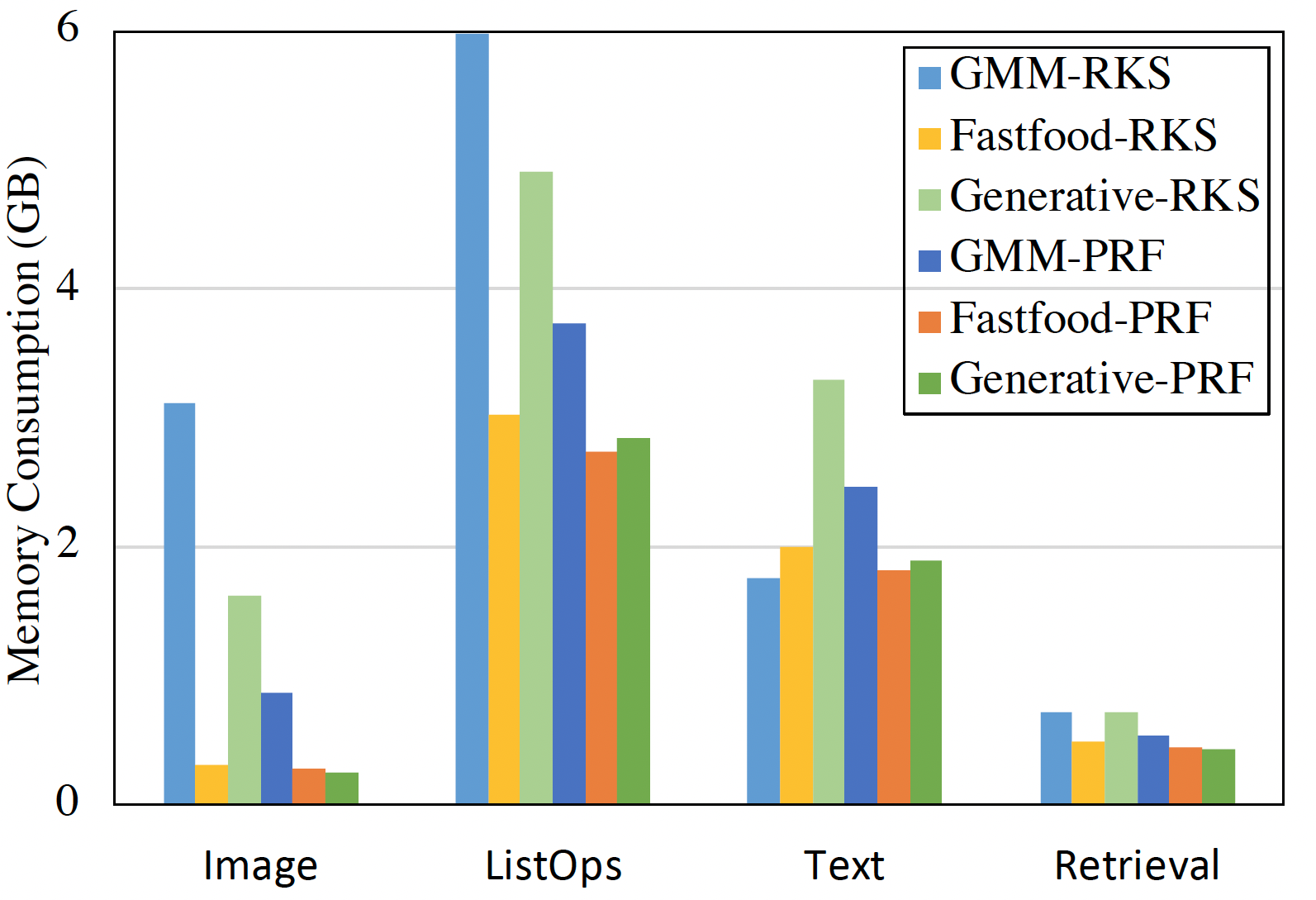}
    \caption{Peak memory used by \kerneltransformer s  across different datasets. }
        \label{fig:lra-memory-length_diff_data}
\end{figure}

\begin{figure}[h]
    \centering 
    \begin{subfigure}[b]{0.5\columnwidth}
        \includegraphics[width=\textwidth]{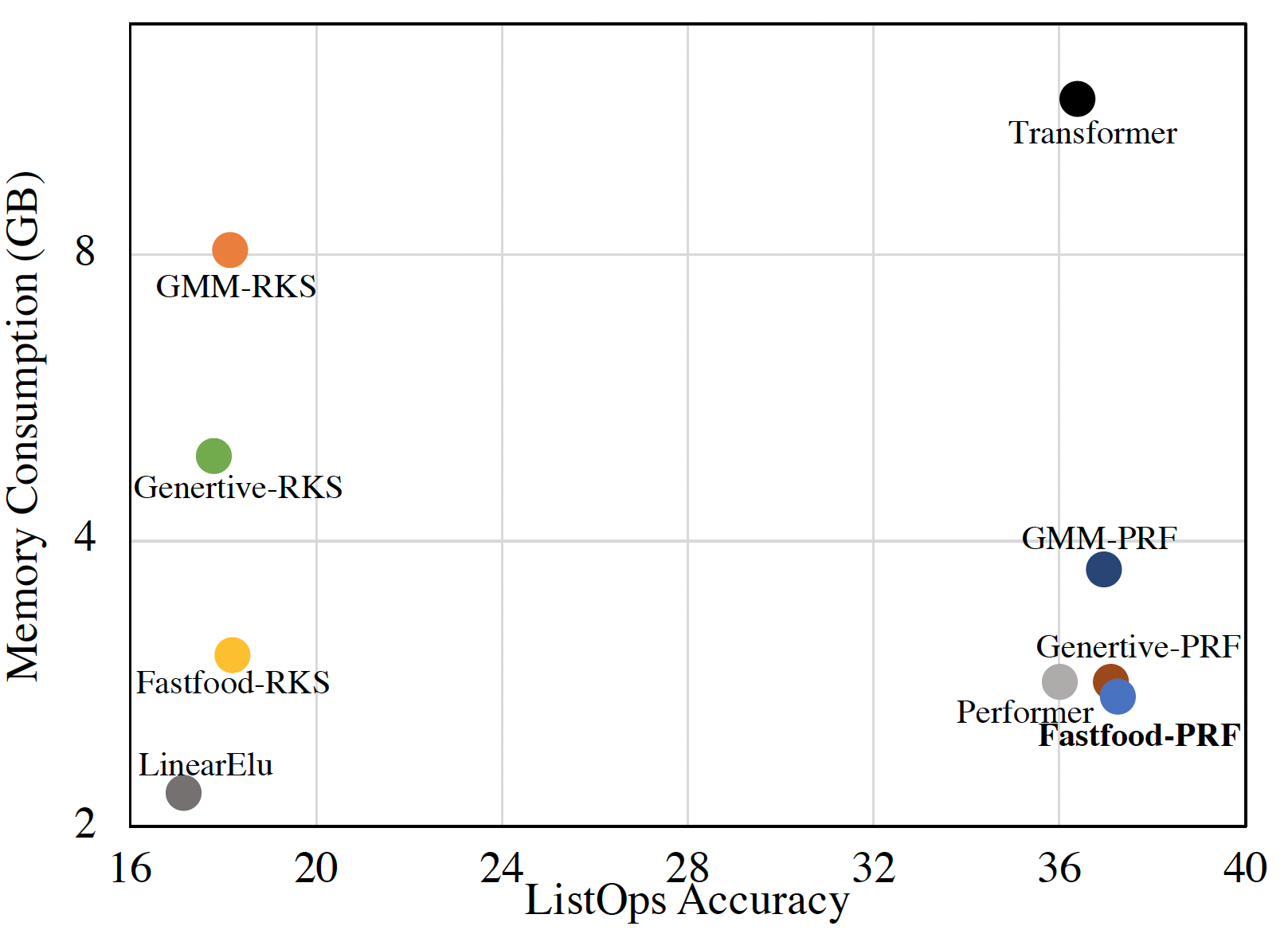}
    \end{subfigure}
    \caption{We demonstrate the peak memory consumption (y-axis) and performance (x-axis) of the various Kernelized Transformer architectures on the ListOps dataset from LRA. Memory usage refers to per device memory usage across each GPU.  
    \label{fig:lra-memory-accuracy-appendix}}
    
\end{figure}

Figure \ref{fig:lra-memory-length_diff_data} shows  how much memory each model uses. Figure \ref{fig:lra-memory-accuracy-appendix} plots this memory usage against the model performance 

\subsection{Correlation of Variance Metrics}
\begin{figure}[h]
    \centering
    \includegraphics[width=0.8\linewidth]{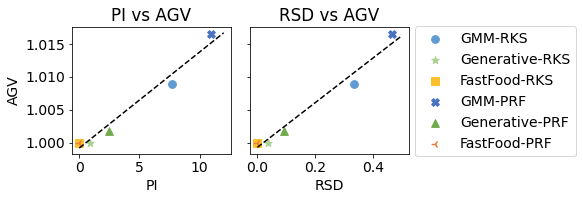}
    \caption{\small Correlation between AGV and the other two variance metrics on the \textbf{Text} task.} \label{fig:analysis-corr}
\end{figure}

\begin{figure}
    \centering
    \includegraphics[width=0.8\textwidth]{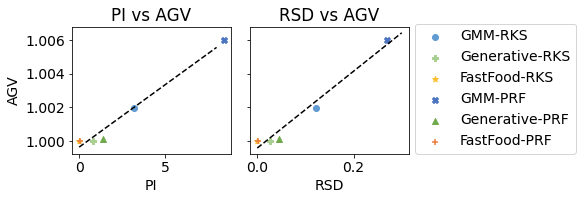}
    \caption{\small Correlation between AGV and the other two variance metrics on the \textbf{Retrieval} task.} .\label{fig:analysis-corr-ret}
\end{figure}

Figures \ref{fig:analysis-corr} and \ref{fig:analysis-corr-ret} show the correlation of Average Gain by Voting with the other twp variance  metrics for the text and retrieval tasks respectively


\subsection{Eigenvalues of Trained Models}
\label{egv}
\begin{table}[]
\centering
\begin{tabular}{|c|c|ccc|}
\hline
Layer              & Head & Minimum  & Maximum  & Mean     \\ \hline
\multirow{4}{*}{1} & 1    & 7.601e-7 & 4.584e-1 & 3.531e-2 \\
                   & 2    & 2.380e-8 & 4.863e-1 & 3.576e-2 \\
                   & 3    & 1.400e-5 & 4.469e-1 & 3.319e-2 \\
                   & 4    & 4.214e-6 & 4.197e-1 & 3.208e-2 \\ \hline
\multirow{4}{*}{2} & 1    & 4.242e-6 & 2.790e-1 & 3.070e-2 \\
                   & 2    & 2.971e-6 & 3.238e-1 & 3.161e-2 \\
                   & 3    & 5.081e-6 & 2.842e-1 & 3.078e-2 \\
                   & 4    & 5.212e-7 & 3.006e-1 & 3.086e-2 \\ \hline
\multirow{4}{*}{3} & 1    & 2.297e-6 & 2.219e-1 & 2.872e-2 \\
                   & 2    & 7.094e-6 & 2.036e-1 & 2.594e-2 \\
                   & 3    & 1.074e-7 & 1.953e-1 & 2.772e-2 \\
                   & 4    & 6.330e-9 & 1.836e-1 & 2.442e-2 \\ \hline
\end{tabular}
\caption{Distribution of Eigenvalues of covariances of \MG\ model for the \textit{Text} task}
\label{tab:egv}
\end{table}
\begin{table}[]
\centering
\begin{tabular}{|c|c|ccc|}
\hline
Layer              & Head & Minimum   & Maximum   & Mean      \\ \hline
\multirow{4}{*}{1} & 1    & 1.363e-5 & 9.660e-2 & 2.440e-2 \\
                   & 2    & 6.659e-5 & 7.900e-2 & 2.200e-2 \\
                   & 3    & 7.241e-6 & 9.050e-2 & 2.810e-2 \\
                   & 4    & 2.760e-5 & 7.790e-2 & 2.630e-2 \\ \hline
\multirow{4}{*}{2} & 1    & 2.068e-5 & 7.520e-2 & 2.430e-2 \\
                   & 2    & 2.386e-5 & 7.580e-2 & 2.400e-2 \\
                   & 3    & 7.256e-6 & 7.400e-2 & 2.370e-2 \\
                   & 4    & 1.911e-4 & 8.080e-2 & 2.680e-2 \\ \hline
\multirow{4}{*}{3} & 1    & 2.871e-5 & 6.710e-2 & 2.570e-2 \\
                   & 2    & 6.117e-6 & 7.370e-2 & 2.190e-2 \\
                   & 3    & 7.830e-6 & 6.600e-2 & 2.010e-2 \\
                   & 4    & 2.186e-5 & 7.620e-2 & 2.810e-2 \\ \hline
\end{tabular}
\caption{Distribution of Eigenvalues of covariances of \PRF\ model for the \textit{Text} task}
\label{tab:egvprf}
\end{table}
\begin{table}[]
\centering
\begin{tabular}{|l|l|lll|}
\hline
Layer              & Head & Minimum   & Maximum   & Mean      \\ \hline
\multirow{4}{*}{1} & 1    & 4.011e-05 & 4.994e-01 & 5.998e-02 \\
                   & 2    & 4.250e-05 & 2.294e-01 & 5.390e-02 \\
                   & 3    & 1.918e-04 & 3.427e-01 & 5.842e-02 \\
                   & 4    & 5.615e-06 & 3.943e-01 & 5.866e-02 \\ \hline
\multirow{4}{*}{2} & 1    & 3.232e-06 & 2.605e-01 & 5.469e-02 \\
                   & 2    & 2.746e-06 & 2.581e-01 & 5.333e-02 \\
                   & 3    & 6.364e-06 & 2.329e-01 & 5.091e-02 \\
                   & 4    & 2.071e-05 & 1.743e-01 & 5.020e-02 \\ \hline
\multirow{4}{*}{3} & 1    & 4.358e-05 & 1.919e-01 & 5.157e-02 \\
                   & 2    & 1.143e-04 & 1.738e-01 & 4.819e-02 \\
                   & 3    & 2.546e-05 & 1.735e-01 & 4.705e-02 \\
                   & 4    & 9.697e-07 & 2.075e-01 & 4.982e-02 \\ \hline
\end{tabular}
\caption{Distribution of Eigenvalues of covariances of \MG\ model for the \textit{Retrieval} task}
\label{tab:egv-ret-mg}
\end{table}
\begin{table}[]
\centering
\begin{tabular}{|l|l|lll|}
\hline
Layer              & Head & Minimum   & Maximum   & Mean      \\ \hline
\multirow{4}{*}{1} & 1    & 8.225e-05 & 2.114e-01 & 4.818e-02 \\
                   & 2    & 5.386e-05 & 1.525e-01 & 4.066e-02 \\
                   & 3    & 9.659e-05 & 1.707e-01 & 5.638e-02 \\
                   & 4    & 1.498e-07 & 1.467e-01 & 4.166e-02 \\ \hline
\multirow{4}{*}{2} & 1    & 1.294e-06 & 9.798e-02 & 3.024e-02 \\
                   & 2    & 9.306e-05 & 1.289e-01 & 4.334e-02 \\
                   & 3    & 4.300e-06 & 1.865e-01 & 4.846e-02 \\
                   & 4    & 2.418e-05 & 1.047e-01 & 4.458e-02 \\ \hline
\multirow{4}{*}{3} & 1    & 2.466e-08 & 1.133e-01 & 3.987e-02 \\
                   & 2    & 2.291e-04 & 1.429e-01 & 5.099e-02 \\
                   & 3    & 3.997e-05 & 1.134e-01 & 3.380e-02 \\
                   & 4    & 6.653e-05 & 1.129e-01 & 2.827e-02 \\ \hline
\end{tabular}
\caption{Distribution of Eigenvalues of covariances of \PRF\ model for the \textit{Retrieval} task}
\label{tab:egv-ret-prf}
\end{table}

\section{Ablation Studies}\label{Ablations}

\subsection{FastFood Attention}
In the main paper, we use FastFood-SGB, which has all the diagonal matrices learnable. However, $B$ and $G$ matrices have a very special structure (their elements being drawn from $Bernoulli_{\{-1,1\}}(0.5)$  and $\mathcal{N}(0,1)$ respectively), which is lost if we make them learnable. Therefore, it makes sense to have \textit{FastFood-S}, which only has $S$ learnable. Finally, we can also have everything fixed, giving us the basic \textit{FastFood} version. The results of these two versions, along with the original FastFood-SGB kernel on the \glue\ benchmaark are summarised in Table \ref{tab:ff}. As one can see, FastFood-SGB is either the best or close to it except for WNLI and CoLA, therefore we choose to use this version for our main analysis.\\
\begin{table}[!h]
\centering
\resizebox{14cm}{!}{
\begin{tabular}{|c|ccccccccccc|}
\hline
\textbf{Dataset}                                                & \textbf{\begin{tabular}[c]{@{}c@{}}SST2\\ (acc)\end{tabular}} & \textbf{\begin{tabular}[c]{@{}c@{}}MRPC\\ (acc)\end{tabular}} & \textbf{\begin{tabular}[c]{@{}c@{}}MRPC\\ (f1)\end{tabular}} & \textbf{\begin{tabular}[c]{@{}c@{}}QQP\\ (acc)\end{tabular}} & \textbf{\begin{tabular}[c]{@{}c@{}}QQP\\ (f1)\end{tabular}} & \textbf{\begin{tabular}[c]{@{}c@{}}MNLI\\ (mat)\end{tabular}} & \textbf{\begin{tabular}[c]{@{}c@{}}MNLI\\ (mis)\end{tabular}} & \textbf{\begin{tabular}[c]{@{}c@{}}QNLI\\ (acc)\end{tabular}} & \textbf{\begin{tabular}[c]{@{}c@{}}WNLI\\ (acc)\end{tabular}} & \textbf{\begin{tabular}[c]{@{}c@{}}RTE\\ (acc)\end{tabular}} & \textbf{\begin{tabular}[c]{@{}c@{}}CoLA\\ (MCorr)\end{tabular}} \\ \hline
\textbf{FastFood}                                               & 0.814                                                         & 0.713                                                         & 0.820                                                        & 0.811                                                        & 0.738                                                       & 0.571                                                         & 0.568                                                         & 0.629                                                         & 0.634                                                         & 0.563                                                        & 0.152                                                           \\
\textbf{FastFood-S}                                             & 0.807                                                         & 0.706                                                         & 0.822                                                        & 0.810                                                        & 0.741                                                       & 0.571                                                         & 0.571                                                         & 0.642                                                         & 0.606                                                         & 0.570                                                        & 0.101                                                           \\
\textbf{\begin{tabular}[c]{@{}c@{}}FastFood\\ SGB\end{tabular}} & 0.828                                                         & 0.707                                                         & 0.820                                                        & 0.810                                                        & 0.739                                                       & 0.569                                                         & 0.572                                                         & 0.638                                                         & 0.592                                                         & 0.563                                                        & 0.129                                                           \\ \hline
\end{tabular}
}
\caption{Ablation studies using FastFood variants on the \glue\ benchmark.}
\label{tab:ff}
\end{table}
\section{Sparsity Synthetic Experiment}
\label{sec:task}
\subsection{Task Description}
Given a sequence of ordered pairs $(v_i,a_i)$, where $v_i \in \{-1,1\}$ and $a_i \in \{0,1\}$, the task is to output $\sum_{i=0}^Lv_ia_i$. Here, $v_i$ can be seen as the value of a given position, while $a_i$ indicates whether or not we need to attend to that position. The dataset is generated by (pseudo-randomly flipping a coin independently for $a_i$ and $v_i$. The bias in the flip for $a_i$ defines the sparsity of the dataset. The flip for $v_i$ is appropriately biased to ensure that no prefix has an absolute sum of more that $4$. The final prediction is outputted as a 9-way classification over integer values between $-4$ and $4$ (both inclusive). The bias against higher absolute values causes $-4$ and $4$ classes to appear less often. This is balanced out by overgeneration and sampling. 

We generate 3 datasets corresponding to sparsities $0.1$, $0.5$ and $0.9$. Each dataset has $200K$ instances, of sequence length $200$. Of these, we use $80\%$ as the training set and the rest for validation.
\subsection{Model Description}
We use a $3$ layer transformer with $d_{model}=d_{feedforward}=64$ and $d_{query}==d_{value}=16$. The input is encoded as a 3-d many-hot vector $(v_i=1,v_i=-1,a_i)$. this is then passed through an embedding layer and added to learnable position embeddings and passed through the transformer. The final embedding of the $0^{th}$ position is then passed through a hidden layer with $64$ units and then passed on to the final layer for a 9-way softmax. For both linear attention models, we use 64 samples.

All models are trained using AdamW optimizer with $\beta_2=0.98$, $\epsilon=10^{-9}$, weight decay=$0.1$, Learning Rate=$5\times 10^{-6}$ and all other parameters set to default. We use SGD with a batch size of $400$ and cross entropy loss.
\subsection{Gradients}
We create checkpoints for the \MG\ and \PRF\ models when they first achieve validation accuracies of 20\%, 40\%, 60\% and 80\%. For each of these checkpoints, we pass the first $50$ validation datapoints and record the gradients on the classifier layer. This process is repeated $50$ times. Thereafter, we calculate the mean and standard deviation of gradients to each neuron. To avoid cancellation of opposite signs, the mean is calculated over absolute values. The final reported numbers are averages over the $64$ neurons. The results are shown in Table \ref{tab:spar-grad}

\begin{table}[]
\centering
\begin{tabular}{|c|c|cc|cc|cc|}
\hline
\multirow{2}{*}{\begin{tabular}[c]{@{}c@{}}Dataset\\ Sparsity\end{tabular}} &
  \multirow{2}{*}{\begin{tabular}[c]{@{}c@{}}Checkpoint\\ Acc.(\%)\end{tabular}} &
  \multicolumn{2}{c|}{\MG} &
  \multicolumn{2}{c|}{\PRF} &
  \multicolumn{2}{c|}{\MG/\PRF} \\ \cline{3-8} 
                     &    & Std. Dev. & Abs. Mean & Std. Dev. & Abs. Mean & Std. Dev. & Abs. Mean \\ \hline
\multirow{4}{*}{0.1} & 20 & 0.083064           & 0.144132      & 0.079822           & 0.139948      & 0.960974     & 0.97097       \\
                     & 40 & 0.094528           & 0.16609       & 0.11077            & 0.193444      & 1.171828     & 1.164689      \\
                     & 60 & 0.099791           & 0.176456      & 0.116044           & 0.201462      & 1.162868     & 1.141712      \\
                     & 80 & 0.131252           & 0.228456      & 0.154802           & 0.275594      & 1.179433     & 1.206332      \\ \hline
\multirow{4}{*}{0.5} & 20 & 0.069646           & 0.121306      & 0.434024           & 0.53019       & 6.231835     & 4.370671      \\
                     & 40 & 0.11552            & 0.201468      & 6.432806           & 12.17245      & 55.68558     & 60.41891      \\
                     & 60 & 0.142631           & 0.249917      & 3.752173           & 5.886064      & 26.30681     & 23.55206      \\
                     & 80 & 0.166899           & 0.295308      & 22.46943           & 42.14157      & 134.6289     & 142.7036      \\ \hline
\multirow{4}{*}{0.9} & 20 & 0.059153           & 0.102415      & 0.065438           & 0.115866      & 1.106237     & 1.131343      \\
                     & 40 & 0.108255           & 0.186704      & 0.123171           & 0.211465      & 1.137784     & 1.132622      \\
                     & 60 & 0.119243           & 0.206855      & 0.126247           & 0.217067      & 1.058737     & 1.049365      \\
                     & 80 & 0.136816           & 0.240525      & 0.168633           & 0.295976      & 1.232546     & 1.230542      \\ \hline
\end{tabular}
\caption{Means and Variances of gradients received at the classifier layer for the synthetic experiment}
\label{tab:spar-grad}
\end{table}

\end{document}